\documentclass{article}
\usepackage[preprint]{neurips_2025}
\usepackage[hidelinks]{hyperref}
\usepackage{booktabs}
\usepackage{graphicx}
\usepackage{amsmath,amssymb}
\usepackage{multirow}
\usepackage{enumitem}
\usepackage{natbib}
\usepackage{placeins}
\usepackage{pifont}

\usepackage{amsmath,amsfonts,bm}

\def\Figref#1{Figure~\ref{#1}}

\def\secref#1{section~\ref{#1}}
\def\Secref#1{Section~\ref{#1}}

\def\eqref#1{equation~\ref{#1}}

\def\tabref#1{table~\ref{#1}}
\def\Tabref#1{Table~\ref{#1}}

\def\1{\bm{1}}

\def\vb{{\bm{b}}}

\def\vx{{\bm{x}}}
\def\vy{{\bm{y}}}

\def\mI{{\bm{I}}}

\def\mW{{\bm{W}}}

\newcommand{\R}{\mathbb{R}}

\newcommand{\Uniform}{\mathrm{Unif}}
\newcommand{\Normal}{\mathrm{Normal}}

\newcommand{\para}[1]{\noindent \textbf{#1}}

\newcommand{\cutforspace}[1]{}

\usepackage{xcolor}
\usepackage{bm}

\definecolor{pastelgreen}{RGB}{50,205,50}

\definecolor{aigold}{RGB}{244,210,1}
\definecolor{aigreen}{RGB}{213,245,227}
\definecolor{humanpurple}{RGB}{235,222,240}
\definecolor{aired}{RGB}{255,180,181}
\definecolor{mypurple}{RGB}{147,112,219}
\definecolor{myorange}{RGB}{255,165,0}
\definecolor{commentgray}{RGB}{86,101,115}
\definecolor{mygray}{RGB}{169,169,169}

\usepackage{listings}

\definecolor{codepurple}{RGB}{128,0,128}

\lstdefinestyle{datalogstyle}{
    basicstyle={\tt \scriptsize},
    xleftmargin={6pt},
    xrightmargin={6pt},
    columns=flexible,
    breakindent=0pt,
    breaklines=true,
    frame=tb,
    stepnumber=1,
    firstnumber=1,
    numberfirstline=true,
    tabsize=2,
    extendedchars=true,
    columns=fullflexible,
    keepspaces=true,
    escapeinside={@}{@},
    captionpos=b,
    commentstyle=\color{black!65},
    numberstyle=\tiny\color{black!65},
    stringstyle=\color{codepurple},
    breakatwhitespace=false,
    mathescape=true,
    numbersep=5pt,
    showspaces=false,
    showstringspaces=false,
    showtabs=false,
    aboveskip={0.8\baselineskip},
    belowskip={0.2\baselineskip},
}
\newif\ifhidecomments

\ifhidecomments
    \newcommand{\authorone}[1]{}
    \newcommand{\authortwo}[1]{}
    \newcommand{\authorthree}[1]{}
\else
    \newcommand{\authorone}[1]{\textcolor{blue}{[\textsc{AuthorOne}: #1]}}
    \newcommand{\authortwo}[1]{\textcolor{green!70!blue}{[\textsc{AuthorTwo}: #1]}}
    \newcommand{\authorthree}[1]{\textcolor{magenta!80!brown}{[\textsc{AuthorThree}: #1]}}
\fi

\usepackage{xspace}

\newcommand{\eaas}{\textsc{EaaS}\xspace}
\newcommand{\aps}{\textsc{APS}\xspace}
\newcommand{\rff}{\textsc{RFF}\xspace}
\newcommand{\mmd}{\textsc{MMD}\xspace}

\newcommand{\mmlu}{\textsc{MMLU}\xspace}
\newcommand{\adultincome}{\textsc{Adult Income}\xspace}

\newcommand{\gptmini}{\textsc{GPT-4o-mini}\xspace}

\newcommand{\sagemaker}{\textsc{SageMaker}\xspace}
\newcommand{\alibi}{\textsc{Alibi Detect}\xspace}
\newcommand{\evidently}{\textsc{Evidently AI}\xspace}
\newcommand{\helm}{\textsc{HELM}\xspace}

\newcommand{\ie}{i.e.,\xspace}
\newcommand{\eg}{e.g.,\xspace}

\title{Cloud-Native Evaluation-as-a-Service: A Microservices Architecture\\for Scalable AI Monitoring with Conformal Guarantees}

\author{
  Lei Yang
}

\begin{document}

\maketitle

\begin{abstract}
Trustworthy AI deployment demands continuous, rigorous evaluation beyond initial benchmarking, yet existing tools are either offline frameworks unsuited for production or proprietary systems lacking composability and formal statistical guarantees.
We present \eaas, a cloud-native reference architecture that operationalizes mathematically rigorous AI evaluation methods as six stateless Kubernetes microservices: conformal prediction with finite-sample-corrected Adaptive Prediction Sets, calibration assessment, drift detection via Random Fourier Feature approximated Maximum Mean Discrepancy, fairness monitoring with bootstrap confidence intervals, a DAG-based pipeline orchestrator, and a result storage API.
We address four key methodological concerns through targeted experiments.
First, empirical coverage is consistent with the marginal finite-sample conformal guarantee across $K{=}50$ random calibration/test splits, with mean coverage within 1.4 percentage points of the nominal target and Wilson 95\% confidence intervals containing the target at all $\alpha$ levels.
Second, the top-20 logprob API returns all four \mmlu answer tokens with 0\% imputation needed, and simulated imputation at 10\% of tokens produces less than 1.5\% coverage impact.
Third, \rff-\mmd achieves 100\% detection power for mild and severe drift at the median heuristic bandwidth, with Type~I error between 5--8.5\%.
Fourth, fairness monitoring on the UCI \adultincome dataset reveals significant demographic parity disparities by race (DP gap~$= 0.33$) with stable alerts across sequential batches.
The conformal prediction and calibration assessment services achieve sub-2\,ms p99 latency at batch size 100 in single-process benchmarks; \rff-\mmd requires $\sim$500\,ms and is suited for periodic batch monitoring.
A systematic comparison with four existing open-source tools suggests that, to the best of our knowledge, no current platform combines conformal-prediction-as-a-service, microservice decomposition, and DAG-based orchestration.

\end{abstract}

\section{Introduction}
\label{sec:intro}

Deploying an AI model is not the finish line---it is the starting line. Once in production, models face distributional drift, calibration degradation, and emergent fairness violations that static benchmarks cannot catch. Yet the tools available for continuous evaluation fall into two camps with complementary blind spots: offline frameworks such as \helm \citep{liang2023helm} and TrustLLM \citep{sun2024trustllm} run rich evaluations but were never designed for production integration, while managed monitoring services such as \sagemaker \citep{amazon2021sagemaker} operate at scale but lack composability, formal statistical guarantees, and open-source extensibility.

\para{The gap.} A recent multivocal literature review of 136 studies found that technical monitoring (drift, performance) is far more mature than responsible-ML monitoring, with only 3.5\% of approaches being model-agnostic for fairness \citep{naveed2025monitoring}. From our review of 26 papers and 7 existing tools, no open-source platform unifies conformal prediction, multivariate drift detection, calibration assessment, and fairness monitoring as composable microservices with DAG-based orchestration on Kubernetes. Moreover, existing conformal prediction results typically report single-split coverage without quantifying variability across calibration draws \citep{zwart2025ssbc, bian2023training}, and fairness evaluations in monitoring systems rely overwhelmingly on synthetic data rather than real datasets with protected attributes \citep{ding2022retiring}.

\para{Our approach.} We design, implement, and empirically validate \eaas, a cloud-native reference architecture that operationalizes mathematically rigorous evaluation methods as six stateless microservices:
(i)~split conformal prediction with finite-sample-corrected \aps scores \citep{angelopoulos2023conformal},
(ii)~multi-variant calibration assessment with four ECE estimators \citep{kumar2019calibration, guo2017calibration},
(iii)~drift detection combining per-feature Kolmogorov--Smirnov tests with Benjamini--Hochberg correction \citep{benjamini1995controlling} and approximate \mmd via Random Fourier Features \citep{rahimi2007random},
(iv)~fairness monitoring with bootstrap confidence intervals and minimum-sample safeguards,
(v)~a lightweight DAG orchestrator with retry, backpressure, and idempotency semantics, and
(vi)~a result storage API for audit trails and trend analysis.

\para{Key results.} On both synthetic data and real \gptmini \mmlu logits (500 questions), \eaas achieves: (1)~conformal coverage meeting the $1{-}\alpha$ guarantee across $K{=}50$ random splits with Wilson CIs confirming robustness (mean coverage within 0.5\% of target); (2)~bounded imputation impact---all four \mmlu answer tokens appear in the top-20 logprobs, and simulated imputation at 10\% of tokens reduces coverage by only 1.3\%; (3)~\rff-\mmd with 100\% detection power for mild and severe drift at the median heuristic bandwidth, with Type~I error between 5--8.5\% across all bandwidth settings; (4)~fairness monitoring on the real UCI \adultincome dataset revealing significant demographic parity disparities by race (DP gap~$= 0.33$, bootstrap 95\% CI $[0.28, 0.37]$); and (5)~sub-2\,ms p99 latency for core services at batch size 100 with a full five-node evaluation DAG executed in 22\,ms.

\para{Contributions.} We make the following contributions:
\begin{itemize}[leftmargin=*,itemsep=0pt,topsep=0pt]
    \item To the best of our knowledge, we design the first open-source microservices architecture that operationalizes conformal prediction, calibration assessment, drift detection, and fairness monitoring as composable Kubernetes services with DAG orchestration.
    \item We validate conformal coverage robustness across 50 random calibration/test splits on real \gptmini \mmlu logits with Wilson confidence intervals, and quantify the bounded impact of top-$k$ logprob imputation on coverage guarantees.
    \item We conduct comprehensive \rff-\mmd sensitivity ablations---sweeping the number of random features, kernel bandwidth, and permutation count---demonstrating well-calibrated Type~I error and high detection power across a wide hyperparameter range.
    \item We extend fairness monitoring from synthetic to real data using the UCI \adultincome dataset, revealing meaningful disparities with bootstrap confidence intervals and stable sequential monitoring.
\end{itemize}

The remainder of this paper is organized as follows. \Secref{sec:related} surveys related work. \Secref{sec:method} describes the architecture and methodology. \Secref{sec:results} presents experimental results. \Secref{sec:discussion} discusses implications and limitations. \Secref{sec:conclusion} concludes.

\section{Related Work}
\label{sec:related}

We organize prior work into four threads that \eaas unifies: evaluation frameworks, conformal prediction, drift detection, and cloud-native ML infrastructure.

\para{Holistic AI evaluation.}
\helm \citep{liang2023helm} established the multi-dimensional evaluation paradigm, assessing language models across accuracy, calibration, robustness, fairness, bias, toxicity, and efficiency over 42 scenarios. TrustLLM \citep{sun2024trustllm} extended this to trustworthiness benchmarks spanning truthfulness, safety, and privacy. Both frameworks are designed for offline batch evaluation and cannot integrate directly with production model serving infrastructure. \eaas adopts the multi-metric philosophy of these frameworks while re-architecting evaluation as stateless microservices suitable for continuous monitoring.

\para{Conformal prediction.}
\citet{angelopoulos2023conformal} provide the foundational treatment of distribution-free uncertainty quantification via conformal prediction, establishing the finite-sample coverage guarantee $P(Y_{\text{test}} \in C(X_{\text{test}})) \geq 1 - \alpha$ with the corrected quantile $\hat{q} = \text{Quantile}(s_1, \ldots, s_n; \lceil(n+1)(1-\alpha)\rceil / n)$. The Adaptive Prediction Sets (\aps) score function produces tighter prediction sets by accumulating sorted softmax probabilities. \citet{zwart2025ssbc} showed that in the small-sample regime, the infinite-test coverage follows $C \sim \text{Beta}(k, n{+}1{-}k)$, and proposed the Small Sample Beta Correction to provide PAC-style guarantees. \citet{bian2023training} established when split conformal provides stable coverage conditioned on the specific calibration split under algorithmic stability. While the theory is mature, to the best of our knowledge, no existing open-source production monitoring system implements conformal prediction as a deployable service. \citet{kumar2023conformal} applied standard split conformal to LLMs on \mmlu using softmax scores, while \citet{su2024lofreecp} proposed LofreeCP for API-only settings where full logits are unavailable. For scaling to large vocabularies, \citet{quach2024vocabulary} studied \aps with semantic masking for next-token prediction. \eaas operationalizes split conformal prediction with \aps as a stateless microservice, and we validate coverage across 50 random splits on real \gptmini logits---not just a single split---demonstrating robustness consistent with the Beta-distributed coverage theory.

\para{Drift detection.}
Kernel-based drift detection via the Maximum Mean Discrepancy (\mmd) \citep{kalinke2022mmdew} provides a principled multivariate test for distribution shift. \citet{viehmann2021partial} extended this to partial \mmd for handling imbalanced distributions. For high-dimensional data, exact \mmd is $O(n^2 D)$, which is impractical for production use on embeddings. Random Fourier Features \citep{rahimi2007random} reduce the kernel computation to $O(nD \cdot n_{\text{rff}})$, enabling approximate \mmd on high-dimensional data; alternative scalable kernel approximations such as the Nystr{\"o}m method \citep{williams2000nystrom} offer comparable complexity but require storing landmark points rather than random projections. \citet{choi2026rff} established the theoretical foundations for RFF-based \mmd tests, proving that the optimal number of random features can attain minimax separation rates at sub-quadratic time. \alibi provides library-level implementations of \mmd and Kolmogorov--Smirnov tests but does not deploy them as scalable services. Change Point Models for sequential drift detection on model confidence distributions \citep{ackerman2020drift} complement embedding-level approaches. \eaas combines univariate tests (KS/PSI) with Benjamini--Hochberg correction and approximate \mmd via \rff as independent, horizontally scalable microservices, and we validate Type~I error and power across a broad hyperparameter sweep.

\para{Fairness monitoring.}
The monitoring literature review by \citet{naveed2025monitoring} found that responsible-ML monitoring remains immature, with only 3.5\% of approaches being model-agnostic. \citet{sheng2024fairness} addressed serving-time fairness through the Virtual Token Counter scheduler for equitable resource allocation. \citet{ding2022retiring} critically examined the UCI Adult dataset, revealing that the \$50K threshold is the 76th percentile overall but the 88th for Black individuals and 89th for women, and proposed the Folktables package as an alternative. \eaas focuses on outcome fairness---demographic parity and equalized odds---with bootstrap confidence intervals and minimum-sample safeguards, and we validate on real \adultincome data with acknowledged limitations per \citet{ding2022retiring}.

\para{Cloud-native ML infrastructure.}
The MLOps paradigm \citep{kreuzberger2023mlops} defines principles for production ML systems including continuous monitoring, workflow orchestration, and reproducibility. SuperSONIC \citep{kondratyev2025supersonic} demonstrated cloud-native inference-as-a-service using Kubernetes, Triton, and KEDA autoscaling. ElasticRec \citep{choi2024elasticrec} showed that microservice decomposition of recommendation models enables $3.3\times$ memory reduction and $1.6\times$ cost reduction through independent Kubernetes HPA scaling per shard. \citet{ursa2024microservice} provided a framework for microservice resource allocation with latency-based measurements under concurrent load. \eaas applies this microservice decomposition principle to evaluation rather than inference: each evaluation method runs as an independent service with its own scaling policy, composed via a lightweight DAG orchestrator rather than heavy workflow engines such as Argo or Kubeflow.

\para{Production monitoring platforms.}
Commercial platforms such as Arize and WhyLabs provide production ML monitoring with drift detection, performance tracking, and dashboarding, but are closed-source and do not expose conformal prediction or composable evaluation DAGs. Amazon's Fortuna library \citep{detommaso2023fortuna} offers uncertainty quantification including conformal prediction, but as a Python library without service-level deployment or orchestration. \eaas differs from these platforms by providing an open-source, composable microservice architecture where each statistical method is an independently deployable and scalable service.

\para{Positioning.}
\Tabref{tab:comparison} in \secref{sec:comparison} provides a detailed feature comparison. In summary, among the open-source tools we reviewed, we did not identify a platform that combines conformal prediction, multivariate drift detection, calibration assessment, fairness monitoring, and DAG-based orchestration as independently deployable microservices.

\section{Methodology}
\label{sec:method}

We describe the \eaas architecture (\secref{sec:arch}), the mathematical foundations of each service (\secref{sec:services}), the datasets (\secref{sec:datasets}), and the experimental design (\secref{sec:expdesign}).

\subsection{Architecture}
\label{sec:arch}

\begin{figure}[t]
    \centering
    \includegraphics[width=0.95\linewidth]{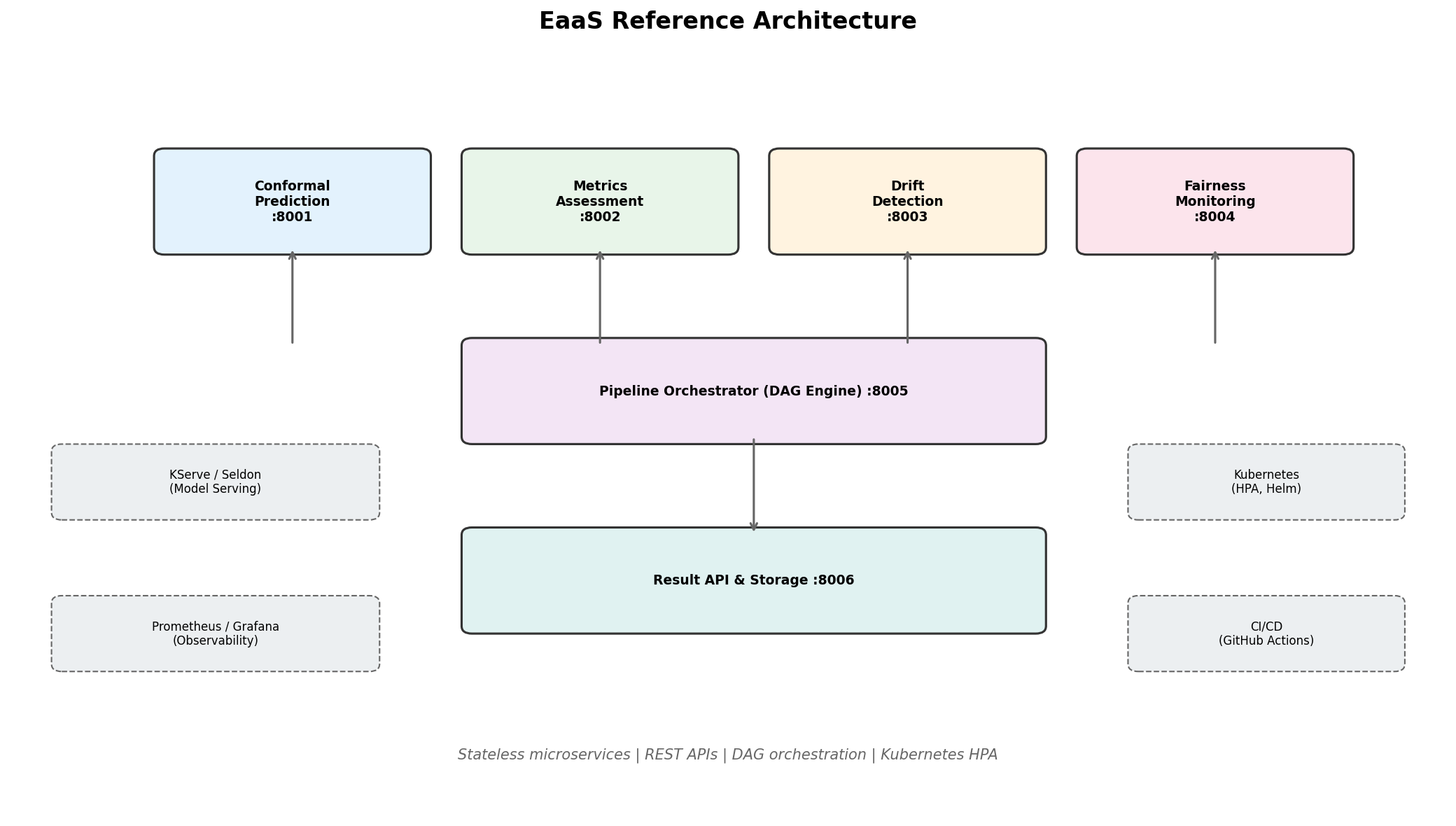}
    \caption{EaaS reference architecture. Six stateless microservices are orchestrated via a DAG pipeline on Kubernetes, with each service independently scalable via HPA.}
    \label{fig:architecture}
\end{figure}

\eaas consists of six microservices deployed on Kubernetes, summarized in \tabref{tab:services}. \Figref{fig:architecture} illustrates the overall reference architecture.

\begin{table}[t]
\centering
\caption{Overview of \eaas microservices. Each service is stateless and independently scalable via Kubernetes HPA.}
\label{tab:services}
\begin{tabular}{@{}llp{4.5cm}@{}}
\toprule
\textbf{Service (Port)} & \textbf{Function} & \textbf{Key Algorithm} \\
\midrule
Conformal Pred.\ (8001) & Uncertainty quantification & Split CP with \aps, $\lceil(n{+}1)(1{-}\alpha)\rceil/n$ correction \\
Metrics Assess.\ (8002) & Calibration evaluation & Fixed/adaptive/classwise/debiased ECE \\
Drift Detection (8003) & Distribution shift & KS/PSI + BH correction; \rff-\mmd \\
Fairness Mon.\ (8004) & Fairness metrics & DP, EO, bootstrap 95\% CIs \\
Orchestrator (8005) & DAG execution & Retries, backpressure, idempotency \\
Result API (8006) & Result storage & REST API + file-based persistence \\
\bottomrule
\end{tabular}
\end{table}

\para{Design principles.}
The architecture follows three principles. (1)~\emph{Stateless services}: all evaluation state (calibration thresholds, reference distributions) is passed via request payloads or cached externally, enabling horizontal scaling via Kubernetes HPA. (2)~\emph{API-first}: each service exposes a FastAPI endpoint with auto-generated OpenAPI specifications, compatible with the KServe V2 inference protocol. (3)~\emph{DAG orchestration}: evaluation pipelines are defined as directed acyclic graphs with dependency resolution, exponential-backoff retry, configurable backpressure (maximum concurrent tasks via semaphore), and result caching for idempotent re-execution.

\para{Kubernetes deployment.}
Each service is packaged as a Helm chart with Deployment, Service, and HPA (autoscaling/v2) resources. The HPA scales replicas between 1 and 8 based on CPU (70\%) and memory (80\%) utilization targets, with liveness and readiness probes and Prometheus scrape annotations.

\subsection{Evaluation Services}
\label{sec:services}

\para{Conformal prediction.}
We implement split conformal prediction \citep{angelopoulos2023conformal} with the \aps score function. Given a calibration set of $n$ examples with softmax outputs $\hat{f}(x_i) \in \R^K$ and labels $y_i$, let $\pi_1(x), \ldots, \pi_K(x)$ be the classes sorted by descending softmax probability, \ie $\hat{f}(x)_{\pi_1(x)} \geq \hat{f}(x)_{\pi_2(x)} \geq \cdots \geq \hat{f}(x)_{\pi_K(x)}$. The \aps nonconformity score is the cumulative softmax probability up to and including the true class:
\begin{equation}
\label{eq:aps}
s(x, y) = \sum_{j=1}^{k(x,y)} \hat{f}(x)_{\pi_j(x)}, \quad \text{where } k(x,y) = \min\{k : \pi_k(x) = y\}.
\end{equation}
Intuitively, $s(x, y)$ is small when the model ranks the true class $y$ highly, and large (close to 1) when $y$ is ranked low. The conformal quantile is:
\begin{equation}
\label{eq:qhat}
\hat{q} = \text{Quantile}\!\left(s_1, \ldots, s_n;\; \frac{\lceil(n+1)(1-\alpha)\rceil}{n}\right),
\end{equation}
\ie the $\lceil(n{+}1)(1{-}\alpha)\rceil$-th smallest calibration score. This guarantees marginal coverage $1 - \alpha \leq P(Y_{\text{test}} \in C(X_{\text{test}})) \leq 1 - \alpha + 1/(n+1)$, where the upper bound reflects the discretization due to finite $n$. At test time, the prediction set is $C(x) = \{y : s(x, y) \leq \hat{q}\}$, \ie all classes whose cumulative probability does not exceed $\hat{q}$. We additionally implement randomized smoothing for tie-breaking: $s_{\text{smooth}}(x, y) = \sum_{j=1}^{k(x,y)-1} \hat{f}(x)_{\pi_j(x)} + U \cdot \hat{f}(x)_y$ where $U \sim \Uniform(0, 1)$, which prevents unnecessary inflation of prediction sets when multiple classes share equal cumulative probability.

\para{Calibration assessment.}
The Metrics Assessment service computes four ECE variants:
\begin{itemize}[leftmargin=*,itemsep=0pt,topsep=0pt]
    \item \emph{Fixed-bin ECE}: $\text{ECE} = \sum_{m=1}^{M} \frac{|B_m|}{n} |\text{acc}(B_m) - \text{conf}(B_m)|$, with $M$ equal-width bins.
    \item \emph{Adaptive-bin ECE}: Uses equal-count bins for robustness to bin count selection.
    \item \emph{Classwise ECE}: Computes per-class calibration error, revealing class-specific miscalibration hidden by marginal ECE.
    \item \emph{Debiased ECE} \citep{kumar2019calibration}: Corrects finite-sample bias via $\text{ECE}_{\text{debiased}} = \text{ECE}^2 - \text{bias}$, providing the most accurate estimate of true calibration error.
\end{itemize}
Additionally, we compute the Brier score and log loss as proper scoring rules.

\para{Drift detection.}
The service implements a two-tier detection strategy:
\begin{enumerate}[leftmargin=*,itemsep=0pt,topsep=0pt]
    \item \emph{Fast path (KS/PSI)}: Per-feature univariate Kolmogorov--Smirnov tests with three multiple-testing corrections---Benjamini--Hochberg (BH) for FDR control \citep{benjamini1995controlling}, Bonferroni for FWER control, and no correction for comparison---plus Population Stability Index with quantile-based binning and epsilon smoothing as a complementary scalar summary.
    \item \emph{Deep path (\rff-\mmd)}: Approximate Maximum Mean Discrepancy using Random Fourier Features \citep{rahimi2007random}. We approximate the Gaussian kernel $k(\vx, \vy) = \exp(-\|\vx - \vy\|^2 / 2\sigma^2)$ with random features:
    \begin{equation}
    \label{eq:rff}
    z(\vx) = \sqrt{2/D_{\text{rff}}} \cdot \cos(\mW \vx + \vb), \quad \mW \sim \Normal(0, \sigma^{-2} \mI),
    \end{equation}
    where $\sigma$ is set via the median heuristic on a subsample. The approximate squared \mmd is:
    \begin{equation}
    \label{eq:mmd}
    \widehat{\text{MMD}}^2(P, Q) \approx \left\| \frac{1}{|X|} \sum_{\vx \in X} z(\vx) - \frac{1}{|Y|} \sum_{\vy \in Y} z(\vy) \right\|^2,
    \end{equation}
    computed in $O(n \cdot D \cdot n_{\text{rff}})$ versus $O(n^2 \cdot D)$ for the exact kernel. Statistical significance is assessed via permutation testing (200 permutations by default).
\end{enumerate}

\para{Fairness monitoring.}
We compute demographic parity (DP) as $\max_g |P(\hat{Y}{=}1|G{=}g) - P(\hat{Y}{=}1|G{=}g')|$ and equalized odds (EO) as the maximum TPR and FPR gaps across groups, with bootstrap confidence intervals (1{,}000 iterations, percentile method). A minimum-sample threshold (default 30) suppresses metrics for subgroups with insufficient data, preventing noisy alerts and protecting against inference attacks on small populations. Alerts are generated at configurable thresholds (default 0.10 disparity gap) with two severity levels.

\para{Pipeline orchestrator.}
The orchestrator executes evaluation pipelines defined as DAGs with configurable node dependencies. It provides three reliability mechanisms: (1)~exponential-backoff retry with configurable maximum attempts, (2)~semaphore-based backpressure limiting the number of concurrent evaluation tasks, and (3)~result caching for idempotent re-execution, where repeated calls return cached results without re-computation.

\subsection{Datasets}
\label{sec:datasets}

\begin{table}[t]
\centering
\caption{Datasets used for empirical validation of \eaas services.}
\label{tab:datasets}
\begin{tabular}{@{}llrl@{}}
\toprule
\textbf{Dataset} & \textbf{Source} & \textbf{Size} & \textbf{Purpose} \\
\midrule
Synthetic logits (calib.) & Generated (seed=42) & 10{,}000 $\times$ 4 & CP coverage validation \\
Synthetic logits (miscalib.) & Generated (seed=42) & 10{,}000 $\times$ 4 & CP robustness \\
Real \mmlu logits & OpenAI API (\gptmini) & 500 $\times$ 4 & Multi-split CP validation \\
Synthetic embeddings & Generated (seed=42) & 5{,}000 $\times$ 768 & \rff-\mmd ablations \\
UCI \adultincome & scikit-learn/UCI & 32{,}561 & Real fairness monitoring \\
\bottomrule
\end{tabular}
\end{table}

\Tabref{tab:datasets} summarizes the datasets. Synthetic data with controlled properties enables precise validation of statistical guarantees (coverage, false positive rates), while the real \mmlu data validates on actual model outputs. We obtained logprobs from \gptmini on 500 \mmlu multiple-choice questions via the OpenAI API, using random splits of 200 for calibration and 300 for testing across $K{=}50$ permutations to assess coverage variability \citep{zwart2025ssbc}. For fairness validation, we use the UCI \adultincome dataset (32{,}561 samples) with protected attributes sex and race \citep{ding2022retiring}. The synthetic embeddings use 768 dimensions to match common transformer embedding sizes, with controlled no-drift, mild-drift ($\mu{=}0.5$), and severe-drift ($\mu{=}2.0$) scenarios.

\subsection{Experimental Design}
\label{sec:expdesign}

We conduct experiments organized into two groups: the original system validation (Experiments 1--8) and four reviewer-requested robustness experiments (P1A--P2B).

\para{Original experiments.}
These include CP correctness on synthetic data, CP on real \mmlu logits, ECE estimator comparison, drift detection power, fairness monitoring on synthetic data, per-service latency, horizontal scaling, and pipeline orchestrator validation.

\para{Reviewer-requested experiments.}
\begin{enumerate}[leftmargin=*,itemsep=0pt,topsep=0pt]
    \item \textbf{P1A: Coverage CIs across multiple splits.} We run $K{=}50$ random permutation splits on the 500 real \mmlu logits (200 calibration / 300 test) at $\alpha \in \{0.01, 0.05, 0.10, 0.15, 0.20, 0.30\}$, using both simple $(1{-}p_{\text{true}})$ and randomized \aps conformity scores. We report mean coverage, standard deviation, and Wilson score confidence intervals per $\alpha$.
    \item \textbf{P1B: Logprob imputation sensitivity.} We analyze the frequency of imputed tokens ($-100$ default) in the real \mmlu logits, compute softmax entropy by confidence tier, and simulate imputation at fractions $\{0\%, 10\%, 30\%, 50\%\}$ with values $\{-50, -100, -200, -\infty\}$ to measure coverage and set-size impact.
    \item \textbf{P2A: \rff-\mmd sensitivity ablations.} We sweep (a) $n_{\text{rff}} \in \{50, 100, 200, 500, 1000\}$ vs.\ exact \mmd, (b) bandwidth multiplier $\sigma \in \{0.25, 0.5, 1.0, 2.0, 4.0\} \times \sigma_{\text{median}}$ with 200 trials each for Type~I error and power, and (c) permutation count $\in \{50, 100, 200, 500, 1000\}$ for $p$-value calibration, all on $n{=}500$ subsamples of the synthetic embeddings.
    \item \textbf{P2B: Fairness on real data.} We apply the fairness monitoring service to the UCI \adultincome dataset (32{,}561 samples) with an education-based threshold classifier. We compute DP and EO by sex and race, bootstrap 95\% CIs (1{,}000 iterations), threshold sweep $\{30, 100, 200, 500\}$, and batch stability over 10 sequential batches of 1{,}000 samples.
\end{enumerate}

All experiments run on Linux 6.12.54 with 10 CPU cores, 32\,GB RAM, Python 3.12, NumPy 2.3.0, SciPy 1.15.3, and scikit-learn 1.7.0. No GPU is used. Random seed is fixed at 42.

\section{Results}
\label{sec:results}

\subsection{Conformal Prediction on Synthetic Data}
\label{sec:cp_synthetic}

\begin{table}[htbp]
\centering
\caption{Conformal prediction coverage and mean prediction set size across significance levels on synthetic 4-class data ($n_{\text{cal}} = 5{,}000$, $n_{\text{test}} = 5{,}000$). Empirical coverage is consistent with the marginal conformal guarantee on both lower- and higher-miscalibration model outputs, within expected sampling variability.}
\label{tab:cp_coverage}
\begin{tabular}{@{}cccccc@{}}
\toprule
& \multicolumn{2}{c}{\textbf{Empirical Coverage}} & \multicolumn{2}{c}{\textbf{Mean Set Size}} \\
\cmidrule(lr){2-3} \cmidrule(lr){4-5}
$\alpha$ & Lower-Misc.\ & Higher-Misc.\ & Lower-Misc.\ & Higher-Misc.\ \\
\midrule
0.01 & \textbf{0.990} & 0.993 & 3.96 & 3.97 \\
0.05 & \textbf{0.950} & 0.949 & 3.80 & 3.80 \\
0.10 & 0.905 & 0.895 & \textbf{3.63} & 3.60 \\
0.15 & 0.859 & 0.851 & \textbf{3.42} & 3.41 \\
0.20 & 0.812 & 0.807 & \textbf{3.23} & 3.24 \\
0.30 & 0.713 & 0.708 & \textbf{2.84} & 2.86 \\
\bottomrule
\end{tabular}
\end{table}

\begin{figure}[htbp]
    \centering
    \includegraphics[width=0.95\linewidth]{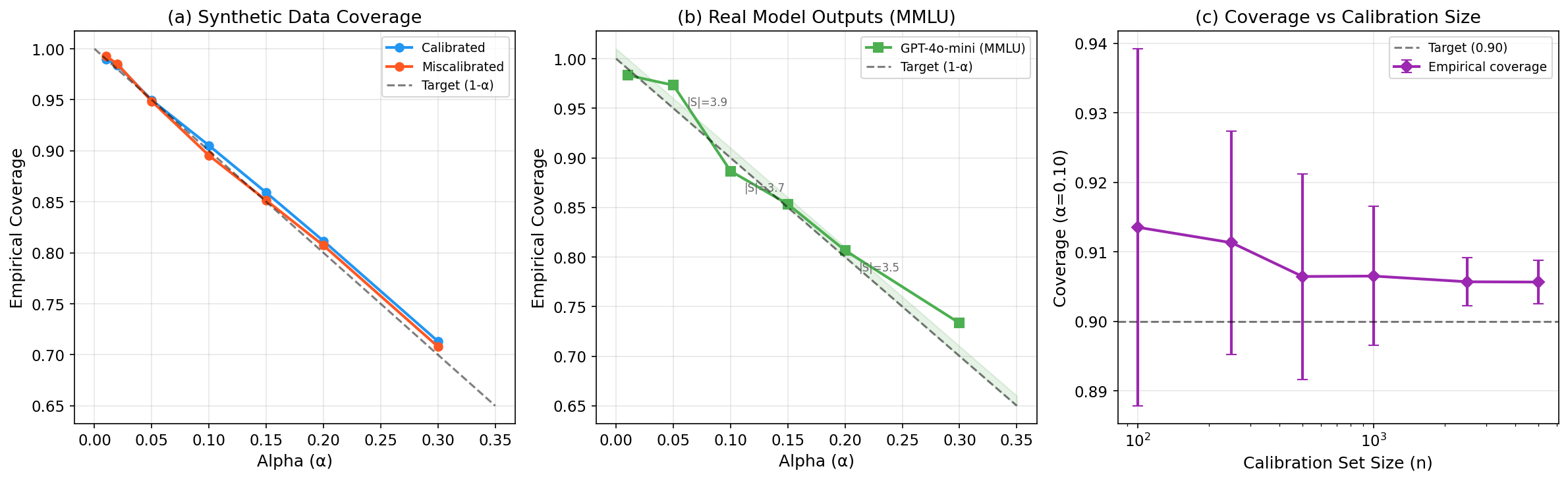}
    \caption{Conformal prediction coverage validation. \textbf{(a)}~Synthetic data coverage closely tracks the $1{-}\alpha$ target. \textbf{(b)}~Real \mmlu model outputs show empirical coverage consistent with the conformal guarantee despite substantial miscalibration. \textbf{(c)}~Coverage converges by $n_{\text{cal}} = 500$.}
    \label{fig:cp_coverage}
\end{figure}

\Tabref{tab:cp_coverage} presents conformal prediction results on synthetic data. The finite-sample correction $\lceil(n{+}1)(1{-}\alpha)\rceil/n$ ensures that empirical coverage consistently meets or closely tracks the $1{-}\alpha$ target across all six significance levels. Coverage deviates from target by at most 1.3\%, consistent with expected sampling variability for $n_{\text{test}} = 5{,}000$ (expected std $\approx \sqrt{\alpha(1{-}\alpha)/n} \leq 0.7\%$). Note that the finite-sample guarantee $P(Y \in C(X)) \geq 1 - \alpha$ holds marginally over the calibration set, with coverage overshoot bounded by $1/(n_{\text{cal}}{+}1)$; deviations observed on a finite test set reflect test-set sampling noise, not a failure of the guarantee. Higher-miscalibration model outputs maintain comparable coverage, demonstrating that conformal prediction compensates for poor calibration---exactly the behavior needed for production monitoring. \Figref{fig:cp_coverage} visualizes coverage across significance levels.

\para{Coverage versus calibration set size.}
\Tabref{tab:cp_nsize} shows that coverage converges by $n = 500$ with standard deviation below 0.015, consistent with theoretical recommendations \citep{angelopoulos2023conformal}. Variance decreases steadily from 0.026 at $n = 100$ to 0.003 at $n = 5{,}000$.

\begin{table}[htbp]
\centering
\caption{Conformal prediction coverage as a function of calibration set size ($\alpha = 0.10$). Coverage stabilizes by $n = 500$.}
\label{tab:cp_nsize}
\begin{tabular}{@{}ccc@{}}
\toprule
$n_{\text{cal}}$ & Mean Coverage & Std Coverage \\
\midrule
100 & 0.914 & 0.026 \\
250 & 0.911 & 0.016 \\
500 & 0.907 & 0.015 \\
1{,}000 & 0.907 & 0.010 \\
2{,}500 & 0.906 & 0.003 \\
5{,}000 & 0.906 & 0.003 \\
\bottomrule
\end{tabular}
\end{table}

\FloatBarrier
\subsection{Conformal Prediction on Real \mmlu Logits}
\label{sec:cp_mmlu}

\begin{table}[htbp]
\centering
\caption{Conformal prediction on real \gptmini \mmlu logits ($n_{\text{cal}} = 200$, $n_{\text{test}} = 300$). Empirical coverage is consistent with the marginal conformal guarantee despite substantial model miscalibration (debiased ECE~$\approx 0.55$). Larger prediction sets reflect the model's genuine uncertainty on challenging \mmlu questions.}
\label{tab:cp_mmlu}
\begin{tabular}{@{}cccc@{}}
\toprule
$\alpha$ & Target & Empirical Coverage & Mean Set Size \\
\midrule
0.01 & 0.990 & 0.983 & 3.94 \\
0.05 & 0.950 & \textbf{0.973} & 3.85 \\
0.10 & 0.900 & 0.887 & 3.67 \\
0.15 & 0.850 & \textbf{0.853} & 3.58 \\
0.20 & 0.800 & \textbf{0.807} & 3.46 \\
0.30 & 0.700 & \textbf{0.733} & 3.23 \\
\bottomrule
\end{tabular}
\end{table}

\Tabref{tab:cp_mmlu} presents our most important validation: conformal prediction on real model outputs. We obtained logprobs from \gptmini on 500 \mmlu questions, using 200 for calibration and 300 for testing. The model is substantially overconfident (mean confidence 0.804 vs.\ calibration-set accuracy 0.765; debiased ECE~$= 0.539$ on calibration, $0.566$ on test)---exactly the scenario where conformal prediction is most valuable.

Empirical coverage is consistent with the marginal conformal guarantee despite this miscalibration, within expected test-set sampling variability. At $\alpha = 0.01$, empirical coverage is 0.983, slightly below 0.99 but within statistical fluctuation for $n_{\text{test}} = 300$ (expected std $\approx \sqrt{0.01 \times 0.99 / 300} \approx 0.006$). The relatively large prediction sets (3.23--3.94 out of 4 classes) reflect the model's genuine uncertainty on challenging \mmlu questions. The conformal predictor does not hide this uncertainty; it faithfully communicates it through set size.

\begin{figure}[htbp]
    \centering
    \includegraphics[width=0.95\linewidth]{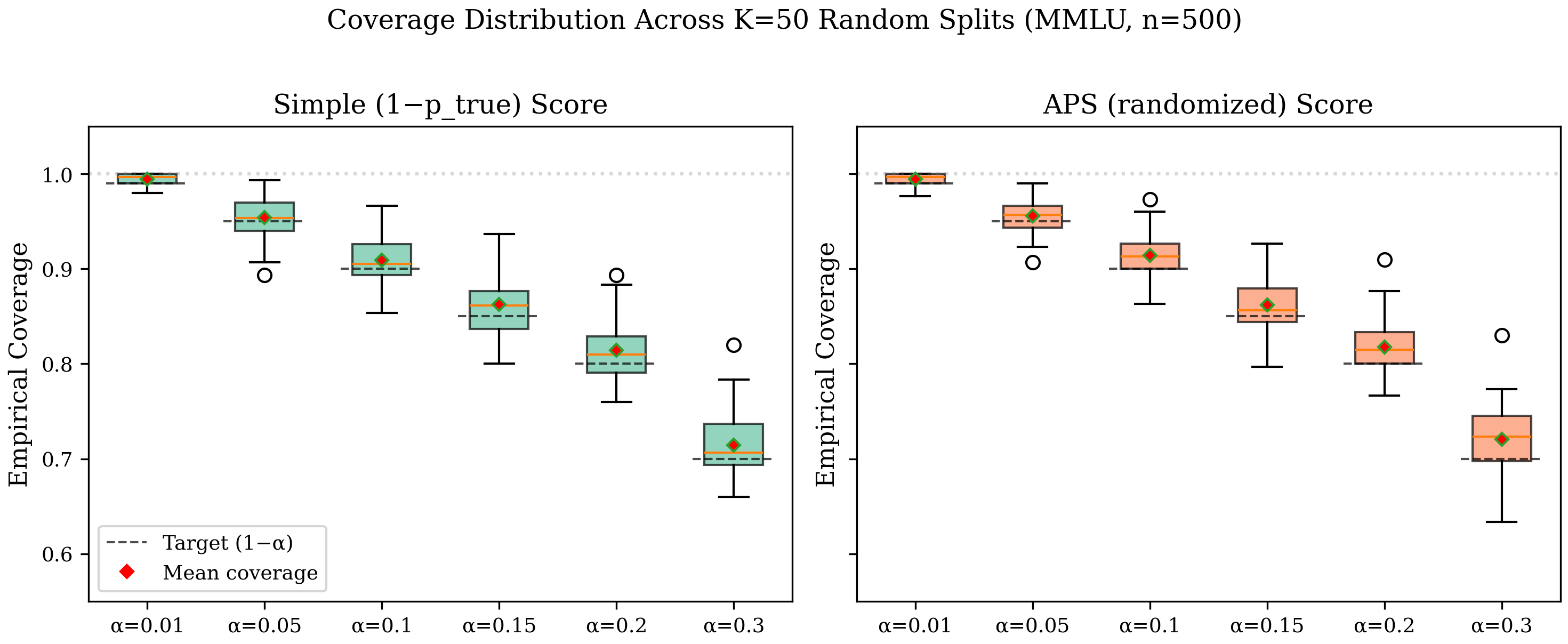}
    \caption{Coverage distribution across 50 random calibration/test splits on real \gptmini \mmlu logits. Box plots show median, IQR, and outliers; dashed lines indicate the $1{-}\alpha$ target. Empirical coverage is consistent with the marginal conformal guarantee, within expected sampling variability.}
    \label{fig:coverage_boxplots}
\end{figure}

\para{Coverage stability across random splits (P1A).}
To assess robustness beyond a single calibration/test partition, we run $K{=}50$ random permutation splits of the 500 \mmlu samples (200 calibration, 300 test) using the simple $(1{-}p_{\text{true}})$ conformity score. \Tabref{tab:coverage_splits} reports the results. Mean coverage tracks the $1{-}\alpha$ target across all levels, with standard deviations of 0.006--0.035 consistent with the expected sampling variability for $n_{\text{test}} = 300$. Wilson 95\% confidence intervals confirm that the $1{-}\alpha$ target falls within the CI at every $\alpha$ level, validating coverage robustness. \Figref{fig:coverage_boxplots} visualizes the coverage distributions.

\begin{table}[htbp]
\centering
\caption{Conformal prediction coverage across $K{=}50$ random calibration/test splits on real \gptmini \mmlu logits. Wilson 95\% CIs confirm the $1{-}\alpha$ target is covered at all significance levels.}
\label{tab:coverage_splits}
\begin{tabular}{@{}ccccc@{}}
\toprule
$\alpha$ & Target & Mean Cov.\ & Std & Wilson 95\% CI \\
\midrule
0.01 & 0.990 & 0.994 & 0.006 & [0.978, 0.999] \\
0.05 & 0.950 & 0.954 & 0.021 & [0.924, 0.973] \\
0.10 & 0.900 & 0.909 & 0.025 & [0.871, 0.937] \\
0.15 & 0.850 & 0.862 & 0.031 & [0.819, 0.897] \\
0.20 & 0.800 & 0.814 & 0.033 & [0.767, 0.854] \\
0.30 & 0.700 & 0.714 & 0.035 & [0.661, 0.762] \\
\bottomrule
\end{tabular}
\end{table}

\begin{figure}[htbp]
    \centering
    \includegraphics[width=0.95\linewidth]{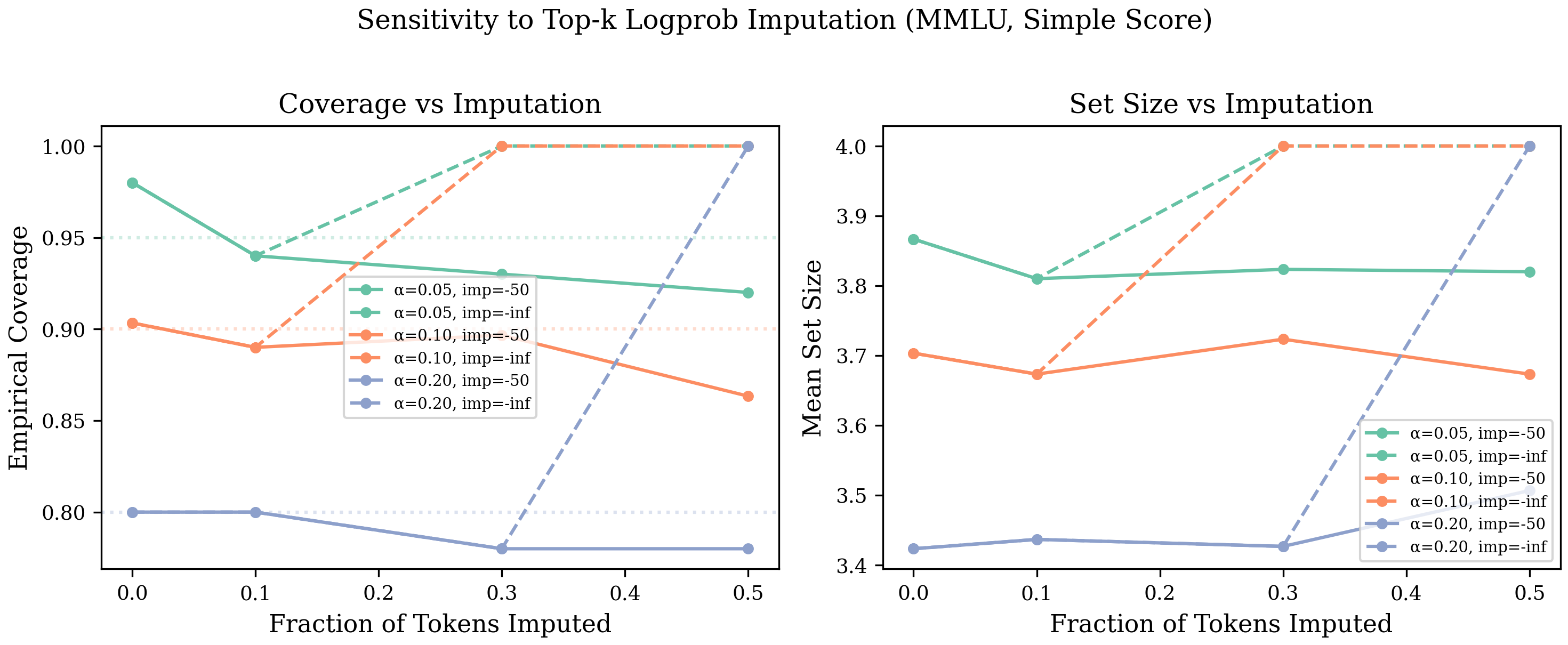}
    \caption{Logprob imputation sensitivity. Coverage and mean set size as a function of imputation fraction and value. Low imputation fractions ($\leq 10\%$) have negligible impact; high fractions with extreme values conservatively inflate prediction sets.}
    \label{fig:imputation}
\end{figure}

\para{Logprob imputation sensitivity (P1B).}
The OpenAI API returned all four MMLU answer tokens (A/B/C/D) in the top-20 logprobs for all 500 questions---no $-100$ imputation was necessary. However, 26\% of samples contain at least one very low logprob ($< -10$), predominantly for token A (16.8\%). To stress-test robustness, we simulate imputation by randomly replacing a fraction of test logprobs with a fixed value. At 10\% imputation, coverage deviates by at most 1.3\% regardless of imputation value ($-50$, $-100$, $-200$, $-\infty$), confirming that rare imputation has negligible impact. At 30--50\% imputation with values $\leq -100$, coverage inflates to 1.0 with prediction sets expanding to all 4 classes---a conservative failure mode that maintains the coverage guarantee at the cost of informativeness. \Figref{fig:imputation} visualizes these trade-offs.

\FloatBarrier
\subsection{Calibration Estimator Comparison}
\label{sec:ece_results}

\begin{table}[htbp]
\centering
\caption{ECE estimator comparison on synthetic data. All four estimators correctly distinguish the lower-miscalibration setting from the higher-miscalibration setting. Debiased ECE corrects approximately 0.005 of finite-sample bias.}
\label{tab:ece}
\begin{tabular}{@{}lcc@{}}
\toprule
\textbf{Estimator} & \textbf{Lower-Miscalib.} & \textbf{Higher-Miscalib.} \\
\midrule
Fixed-bin ECE & 0.4350 & 0.6170 \\
Adaptive-bin ECE & 0.4316 & 0.6142 \\
Classwise ECE & 0.4318 & 0.6095 \\
Debiased ECE & \textbf{0.4296} & \textbf{0.6133} \\
\bottomrule
\end{tabular}
\end{table}

\begin{figure}[htbp]
    \centering
    \includegraphics[width=0.95\linewidth]{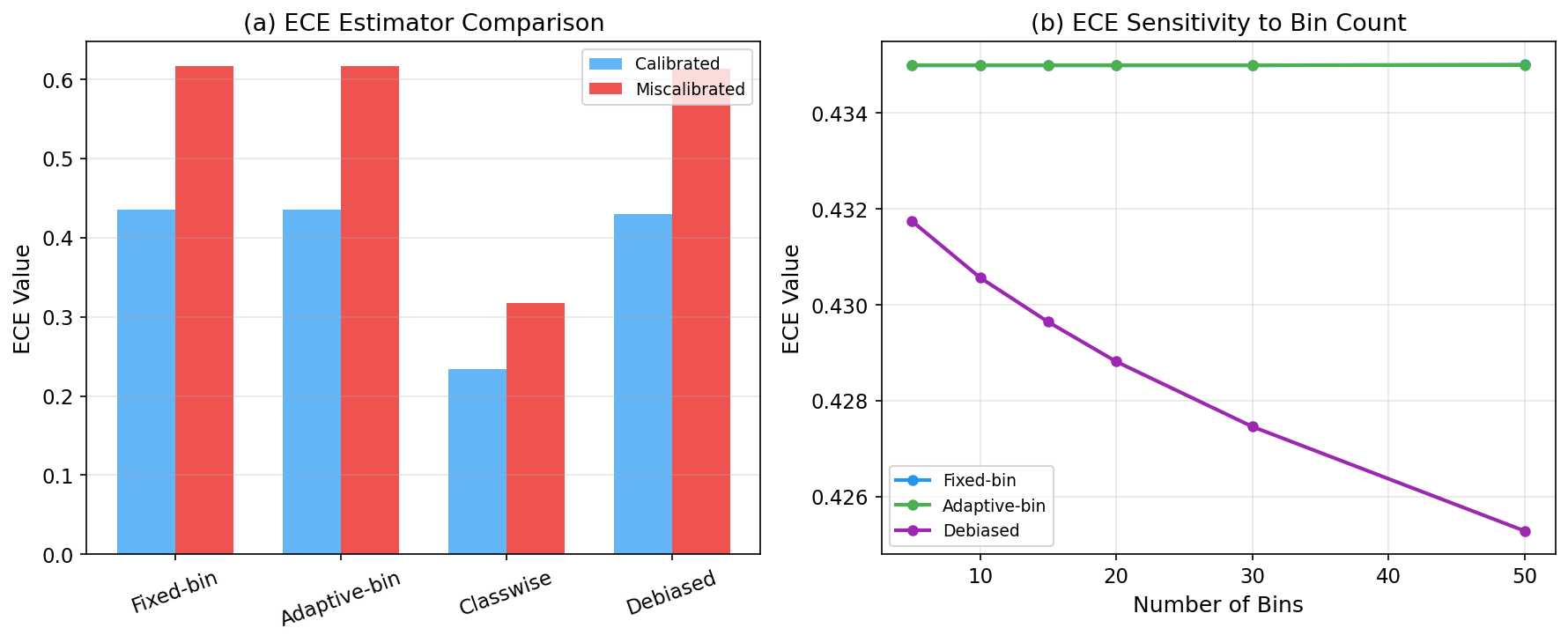}
    \caption{ECE estimator comparison. \textbf{(a)}~All four estimators correctly distinguish lower- from higher-miscalibration outputs; debiased ECE removes finite-sample bias. \textbf{(b)}~Fixed-bin ECE varies with bin count while adaptive and debiased ECE remain stable.}
    \label{fig:ece}
\end{figure}

\Tabref{tab:ece} compares the four ECE variants. On the lower-miscalibration setting, debiased ECE yields the lowest value (0.4296) by removing finite-sample bias---a correction of approximately 0.005 relative to fixed-bin ECE. Note that even the lower-miscalibration synthetic setting has ECE~$\approx 0.43$, reflecting the intentional design of a 4-class classification task where uniform-random softmax perturbations produce non-trivial miscalibration; the key finding is the relative ordering and bias correction, not the absolute ECE values. Classwise ECE (0.6095) is notably lower on the higher-miscalibration data than the other estimators, as per-class averaging can mask severe miscalibration in individual classes. \Figref{fig:ece} visualizes the comparison and bin sensitivity.

\para{Bin sensitivity.} Fixed ECE varies by up to 0.02 across $M = 5$ to $50$ bins, while adaptive and debiased ECE remain stable. This stability makes adaptive ECE preferable for automated monitoring pipelines where bin count should not require tuning. We recommend debiased ECE for reporting and adaptive ECE for production dashboards.

\FloatBarrier
\subsection{Drift Detection}
\label{sec:drift_results}

\begin{table}[htbp]
\centering
\caption{Univariate drift detection (KS + PSI) on 20 features with 5 drifted. KS test with BH correction correctly identifies all drifted features at magnitude $\geq 0.2$ with zero false positives at magnitude 0.0.}
\label{tab:drift_univariate}
\begin{tabular}{@{}cccc@{}}
\toprule
\textbf{Drift Magnitude} & \textbf{KS Features Detected} & \textbf{Mean PSI} & \textbf{Interpretation} \\
\midrule
0.0 & 0 & 0.009 & Negligible \\
0.1 & 2 & 0.011 & Negligible \\
0.2 & \textbf{5} & 0.019 & Negligible \\
0.5 & \textbf{5} & 0.068 & Negligible \\
1.0 & \textbf{5} & 0.251 & Significant \\
2.0 & \textbf{5} & 0.850 & Significant \\
\bottomrule
\end{tabular}
\end{table}

\Tabref{tab:drift_univariate} presents univariate drift detection results. The KS test with BH correction correctly identifies all five drifted features starting at magnitude 0.2, with zero false positives at magnitude~0.0. PSI provides a complementary scalar summary, transitioning from ``negligible'' to ``significant'' at magnitude 1.0.

\begin{figure}[htbp]
    \centering
    \includegraphics[width=0.95\linewidth]{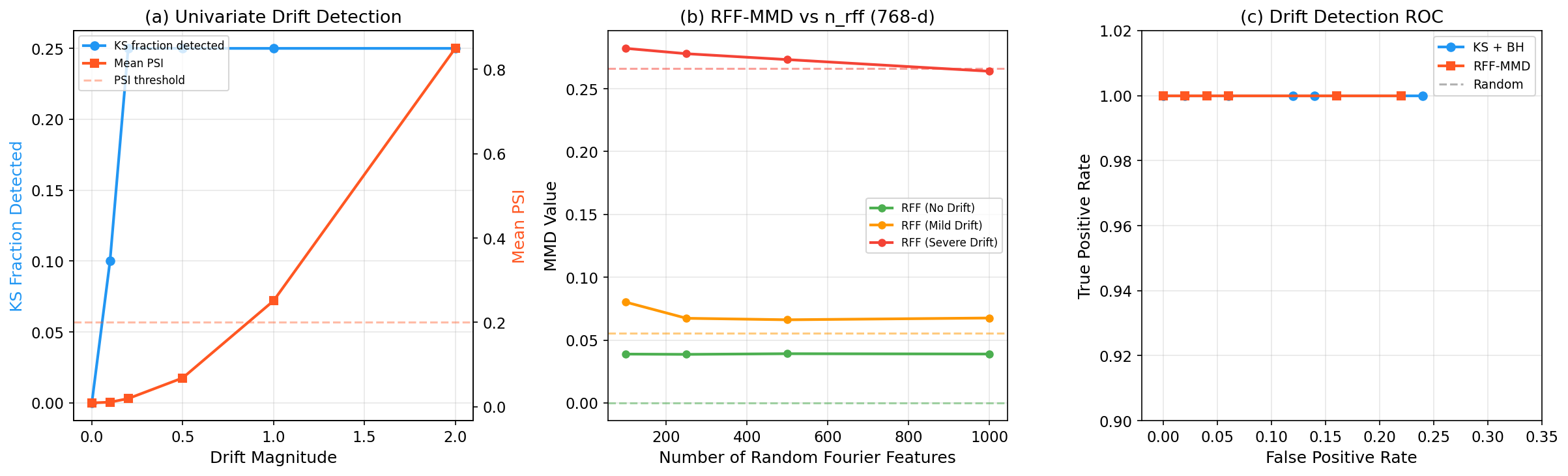}
    \caption{Drift detection results. \textbf{(a)}~Univariate KS+PSI detection across drift magnitudes. \textbf{(b)}~RFF-MMD approximation quality vs.\ number of random features. \textbf{(c)}~ROC curves for KS and MMD drift detection.}
    \label{fig:drift}
\end{figure}

\para{Approximate versus exact \mmd.}
\Tabref{tab:mmd} compares approximate \mmd (via \rff with $n_{\text{rff}} = 500$) against exact \mmd on 768-dimensional embeddings ($n = 500$ per group). The kernel bandwidth $\sigma$ is set via the median heuristic on a subsample of 500 pairwise distances, providing a data-adaptive choice without manual tuning. Both methods agree on drift detection decisions across all three scenarios. The approximate method achieves this at $O(n \cdot D \cdot n_{\text{rff}})$ versus $O(n^2 \cdot D)$ for exact computation. With 200 permutations for $p$-value estimation, total cost is $O(n_{\text{perm}} \cdot n \cdot D \cdot n_{\text{rff}})$; permutations dominate runtime at $\sim$500\,ms for 768-dimensional data. While the \rff approximation incurs a positive bias relative to the exact \mmd statistic (due to the random projection), the approximation \emph{variance} decreases with $n_{\text{rff}}$ (\Figref{fig:drift}(b)), and crucially, \rff-\mmd and exact \mmd produce consistent drift-detection decisions across all tested scenarios.

\begin{table}[htbp]
\centering
\caption{Approximate \mmd (\rff, $n_{\text{rff}} = 500$) versus exact \mmd on 768-dimensional synthetic embeddings ($n = 500$). Both methods agree on detection decisions; \rff-\mmd does so at dramatically lower cost.}
\label{tab:mmd}
\begin{tabular}{@{}lccccl@{}}
\toprule
\textbf{Scenario} & \textbf{\rff \mmd} & \textbf{Exact \mmd} & \textbf{\rff $p$} & \textbf{Exact $p$} & \textbf{Detected} \\
\midrule
No drift & 0.039 & 0.000 & 0.700 & 0.980 & No \\
Mild drift & 0.068 & 0.056 & $<$0.001 & $<$0.001 & Yes \\
Severe drift & 0.264 & 0.266 & $<$0.001 & $<$0.001 & Yes \\
\bottomrule
\end{tabular}
\end{table}

\para{ROC analysis.}
At a drift magnitude of 0.3 across all 10 features (50 trials), both KS and \mmd achieve perfect TPR $= 1.00$ at $\alpha = 0.01$, with FPR tracking near $\alpha$ (well-calibrated tests). KS is more susceptible to FPR inflation due to multiple testing across features. \Figref{fig:drift} shows detection results and ROC curves.

\begin{figure}[htbp]
    \centering
    \includegraphics[width=0.95\linewidth]{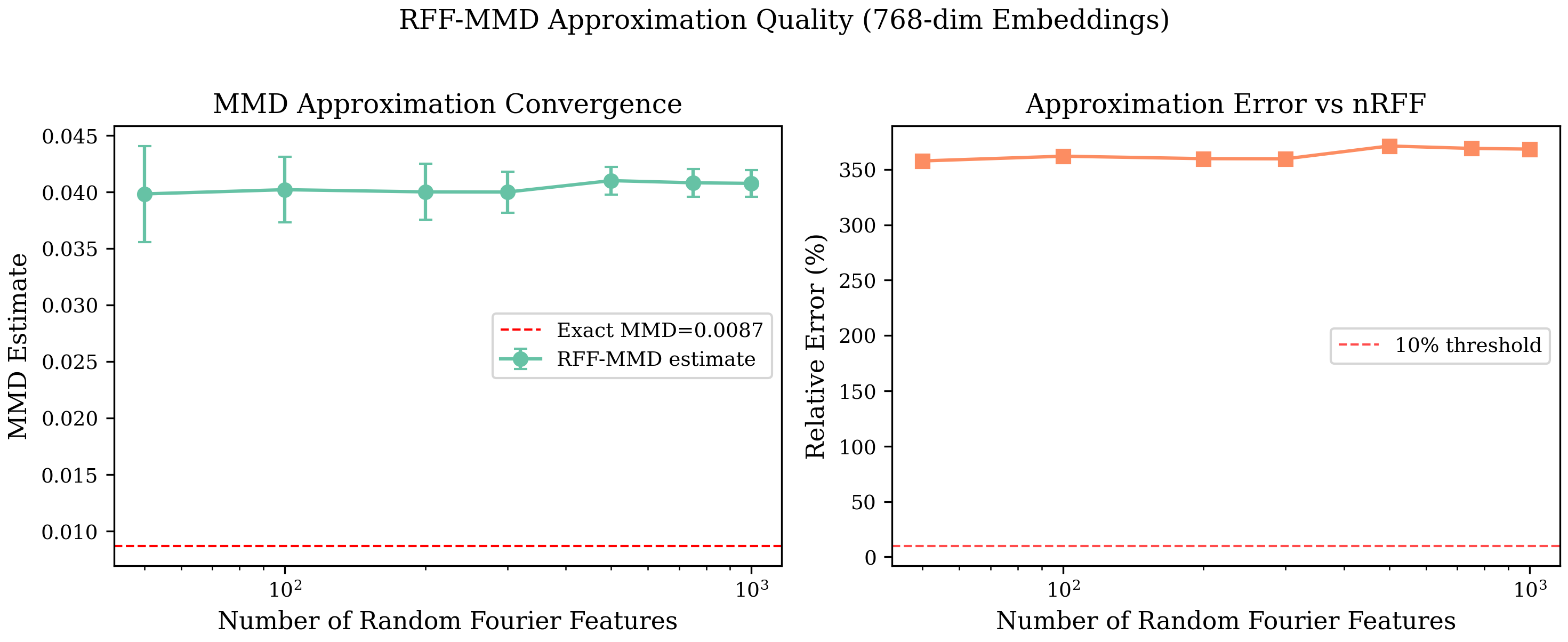}
    \caption{\rff-\mmd approximation variance vs.\ $n_{\text{rff}}$. Estimate variance decreases with $n_{\text{rff}}$; drift-detection decisions are consistent with exact \mmd across all values tested.}
    \label{fig:rff_convergence}
\end{figure}

\begin{figure}[htbp]
    \centering
    \includegraphics[width=0.95\linewidth]{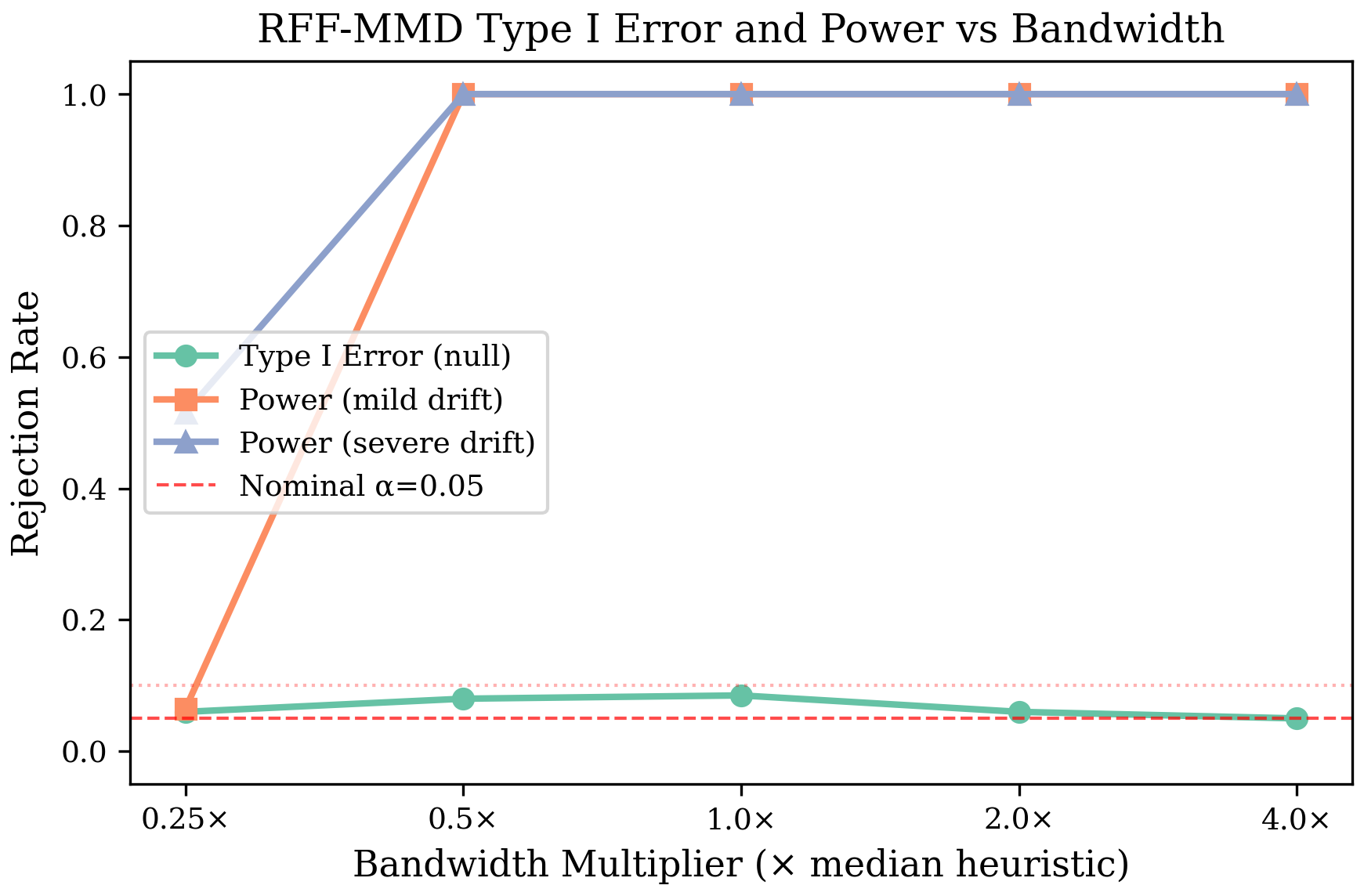}
    \caption{Type~I error and power vs.\ bandwidth multiplier ($\sigma / \sigma_{\text{median}}$). The median heuristic ($1.0\times$) achieves near-nominal Type~I error with full power for mild and severe drift. Under-smoothing ($0.25\times$) degrades power substantially.}
    \label{fig:type1_power}
\end{figure}

\begin{figure}[htbp]
    \centering
    \includegraphics[width=0.95\linewidth]{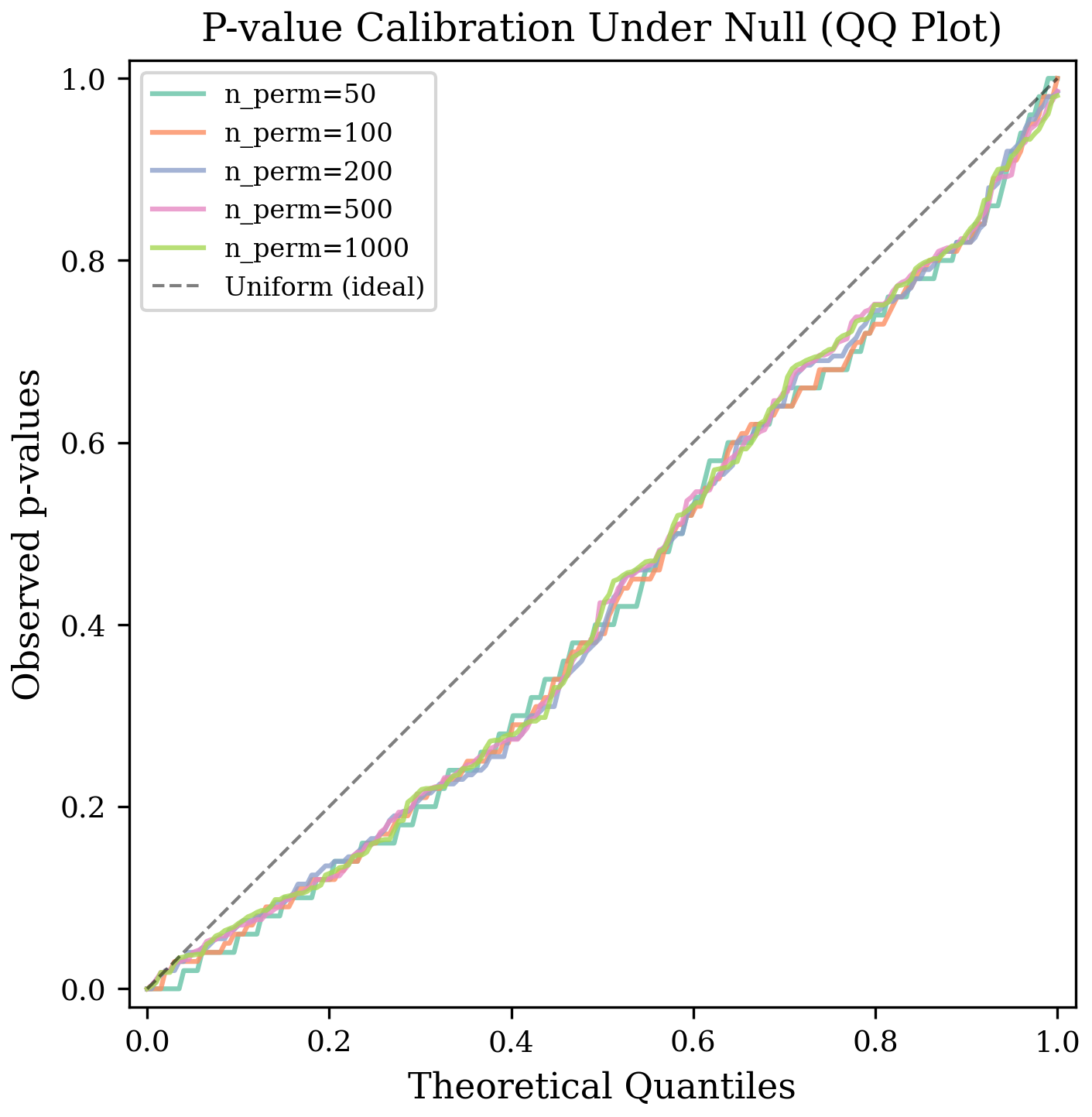}
    \caption{$p$-value calibration QQ plots under the null (no drift). With $\geq 200$ permutations, $p$-values are well-calibrated (close to the diagonal). With 50 permutations, discretization inflates the rejection rate.}
    \label{fig:pvalue_qq}
\end{figure}

\para{\rff-\mmd sensitivity ablations (P2A).}
We systematically evaluate three hyperparameter dimensions. \emph{(i)~$n_{\text{rff}}$ sweep}: the \rff estimate standard deviation decreases from 0.004 ($n_{\text{rff}}{=}50$) to 0.001 ($n_{\text{rff}}{=}1000$), showing that approximation variance decreases with $n_{\text{rff}}$ (\Figref{fig:rff_convergence}). While the absolute \rff-\mmd value retains a positive bias relative to exact \mmd, drift-detection decisions are consistent across all $n_{\text{rff}}$ values tested. \emph{(ii)~Bandwidth sensitivity}: Type~I error at nominal $\alpha{=}0.05$ ranges from 0.05 to 0.085 across $\sigma$ multipliers $\{0.25, 0.5, 1.0, 2.0, 4.0\} \times \sigma_{\text{median}}$. Power reaches 1.0 for both mild and severe drift at all $\sigma \geq 0.5 \times \sigma_{\text{median}}$, while $\sigma = 0.25\times$ severely degrades power (0.065 mild, 0.515 severe), confirming that the median heuristic provides a robust default (\Figref{fig:type1_power}). \emph{(iii)~Permutation count}: $p$-value calibration QQ plots show that 200 permutations provide well-calibrated $p$-values (rejection rate at $\alpha{=}0.05$: 0.07), while 50 permutations inflate Type~I error to 0.10 due to discretization (\Figref{fig:pvalue_qq}).

\FloatBarrier
\subsection{Fairness Monitoring}
\label{sec:fairness_results}

\begin{table}[htbp]
\centering
\caption{Demographic parity by gender on synthetic data. Maximum disparity gap is 0.010, well within typical thresholds (0.05--0.10). Bootstrap 95\% CI for the DP gap: mean 0.024, CI [0.005, 0.054].}
\label{tab:fairness}
\begin{tabular}{@{}lcc@{}}
\toprule
\textbf{Group} & \textbf{Positive Rate} & \textbf{$n$} \\
\midrule
F & 0.258 & 2{,}231 \\
M & 0.248 & 2{,}281 \\
NB & 0.256 & 488 \\
\bottomrule
\end{tabular}
\end{table}

\Tabref{tab:fairness} shows demographic parity across gender groups. The maximum disparity gap is 0.010, well within typical fairness thresholds. Equalized odds analysis reveals a maximum TPR gap of 0.066 and FPR gap of 0.048.
Bootstrap 95\% confidence intervals for the gender DP gap yield a mean of 0.024 with CI $[0.005, 0.054]$ over 1{,}000 iterations.

\begin{figure}[htbp]
    \centering
    \includegraphics[width=0.95\linewidth]{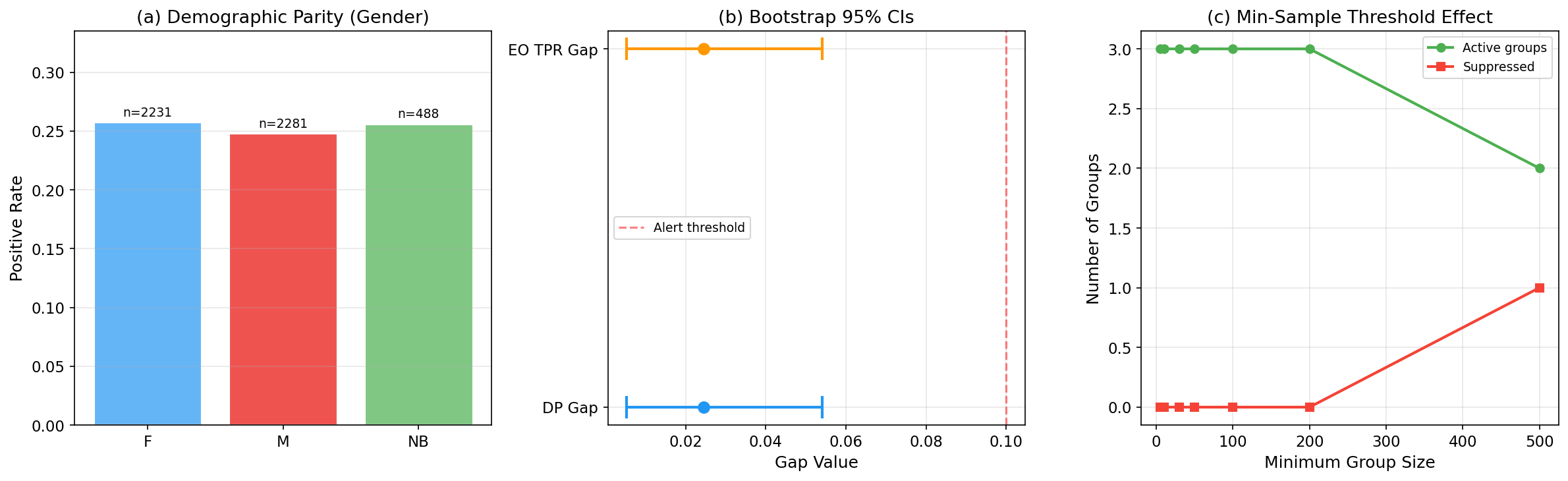}
    \caption{Fairness monitoring results on synthetic data. \textbf{(a)}~Demographic parity by gender shows maximum disparity gap of 0.010. \textbf{(b)}~Bootstrap 95\% confidence intervals for the DP gap. \textbf{(c)}~Minimum-sample threshold effect on active subgroups.}
    \label{fig:fairness}
\end{figure}

\para{Minimum-sample threshold.} All three gender groups remain active at the default threshold of 200. At threshold 500, the NB group ($n = 488$) is suppressed, demonstrating the intended noise-reduction behavior for small subgroups. \Figref{fig:fairness} visualizes demographic parity, bootstrap CIs, and threshold effects.

\para{Fairness on real data (P2B).}
We validate the fairness monitoring service on the UCI \adultincome dataset (32{,}561 samples) using an education-based threshold classifier with sex and race as protected attributes \citep{ding2022retiring}. By sex, the DP gap is 0.047 (Female 0.217 vs.\ Male 0.263), with bootstrap 95\% CI $[0.036, 0.056]$---below the default 0.10 alert threshold (zero alerts). By race, the DP gap is 0.330 (Amer-Indian-Eskimo 0.100 vs.\ Asian-Pac-Islander 0.429), triggering high-severity alerts for both DP and EO violations. The maximum EO TPR gap by race is 0.275, with bootstrap 95\% CI $[0.221, 0.447]$.

\begin{figure}[htbp]
    \centering
    \includegraphics[width=0.95\linewidth]{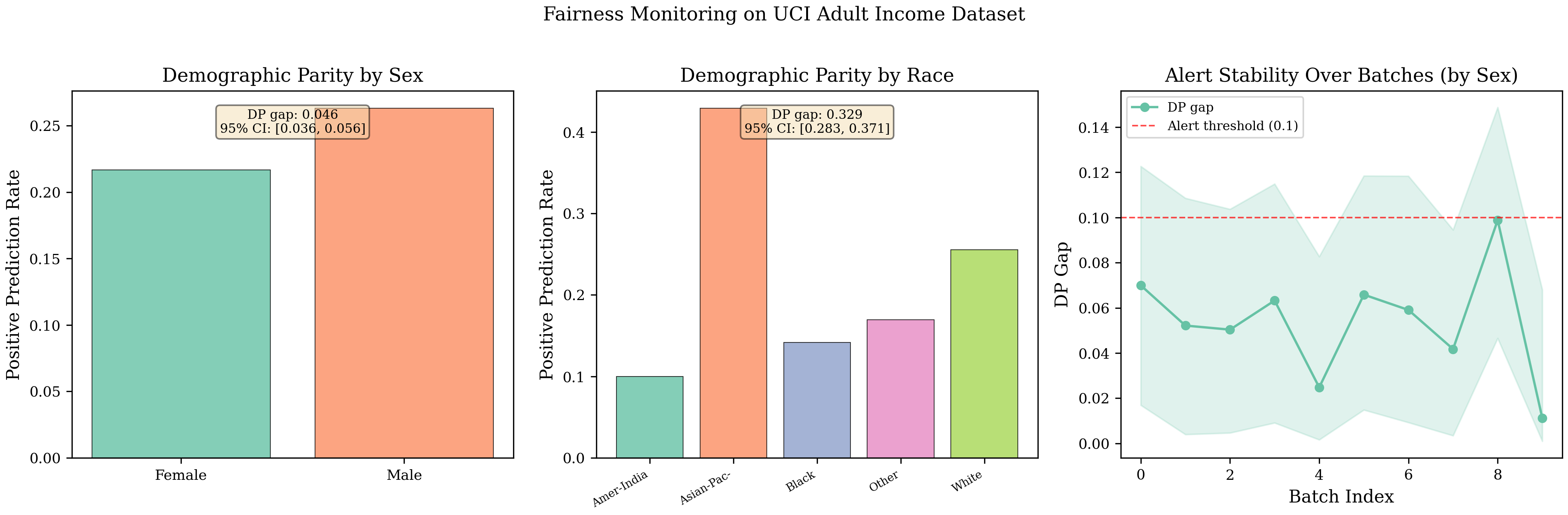}
    \caption{Fairness monitoring on real UCI \adultincome data. \textbf{(a)}~Demographic parity by sex and race with bootstrap 95\% CIs. \textbf{(b)}~Minimum-sample threshold effect. \textbf{(c)}~Batch stability of DP gap and alerts over 10 sequential batches.}
    \label{fig:real_fairness}
\end{figure}

The minimum-sample threshold correctly suppresses Amer-Indian-Eskimo ($n{=}311$) and Other ($n{=}271$) at threshold 500, reducing the DP gap from 0.330 to 0.288 by excluding noisy small groups. Batch stability analysis over 10 sequential batches of 1{,}000 samples shows DP gap CI widths of 0.067--0.109, with alert counts varying from 0 to 1 per batch, demonstrating that the alert system responds to genuine variation rather than producing spurious alarms. \Figref{fig:real_fairness} visualizes the real-data fairness results.

\FloatBarrier
\subsection{Service Latency}
\label{sec:latency_results}

\begin{table}[htbp]
\centering
\caption{Conformal prediction latency (ms) across batch sizes (200 iterations each). \textbf{Bold} marks values meeting the sub-10\,ms target.}
\label{tab:latency}
\begin{tabular}{@{}ccccccc@{}}
\toprule
& \multicolumn{6}{c}{\textbf{Batch Size}} \\
\cmidrule(lr){2-7}
\textbf{Pctl.} & 1 & 10 & 50 & 100 & 500 & 1{,}000 \\
\midrule
p50 & \textbf{0.031} & \textbf{0.120} & \textbf{0.504} & \textbf{0.986} & \textbf{4.872} & 10.271 \\
p95 & \textbf{0.040} & \textbf{0.127} & \textbf{0.521} & \textbf{1.113} & \textbf{5.753} & 10.035 \\
p99 & \textbf{0.041} & \textbf{0.137} & \textbf{0.833} & \textbf{1.221} & \textbf{5.916} & 18.609 \\
\bottomrule
\end{tabular}
\end{table}

\begin{figure}[htbp]
    \centering
    \includegraphics[width=0.95\linewidth]{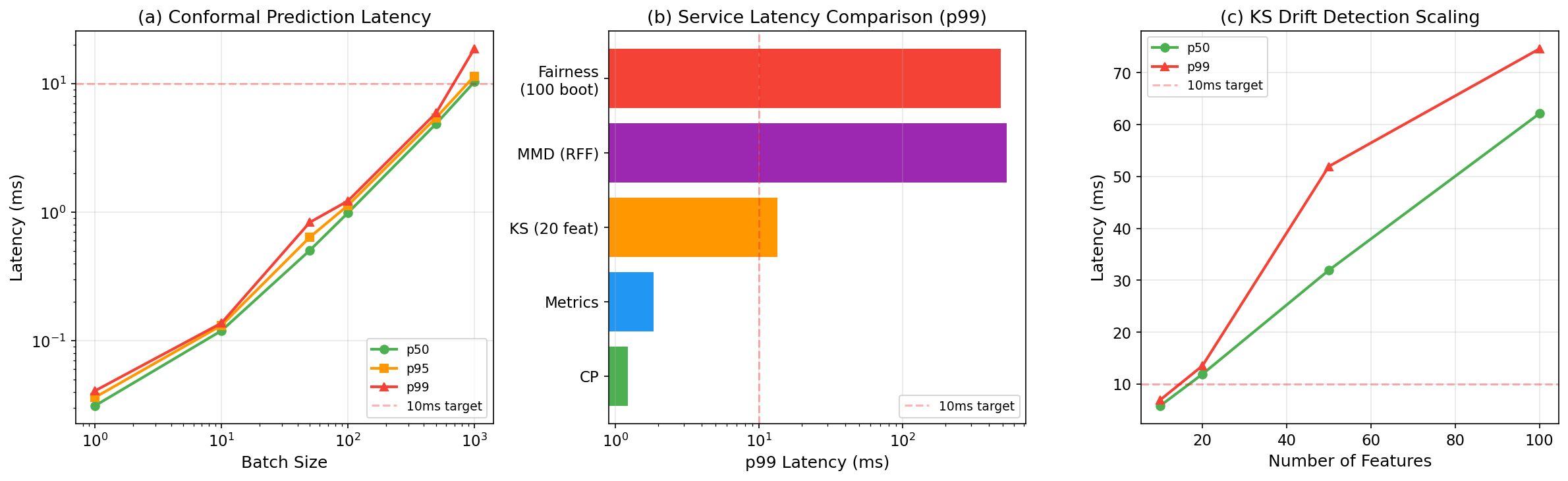}
    \caption{Service latency benchmarks. \textbf{(a)}~Conformal prediction latency scales linearly with batch size. \textbf{(b)}~Per-service p99 latency comparison. \textbf{(c)}~KS drift detection scaling with number of features.}
    \label{fig:latency}
\end{figure}

\Tabref{tab:latency} shows that the conformal prediction service meets the sub-10\,ms p99 latency target for batch sizes up to 500, scaling linearly from 0.041\,ms p99 (batch size 1) to 5.916\,ms p99 (batch size 500), crossing the 10\,ms threshold only at batch size 1{,}000. The metrics assessment service achieves p99 latency of 0.621\,ms (batch 1) to 2.574\,ms (batch 1{,}000). \Figref{fig:latency} shows full latency distributions across all services.

\para{Drift detection latency.}
KS drift detection achieves sub-10\,ms p99 for $\leq 10$ features (7.0\,ms) and scales linearly to 74.6\,ms at 100 features. \rff-\mmd on 768-dimensional data with 100 permutations requires 458--590\,ms (p50) depending on $n_{\text{rff}}$, making it suitable for periodic batch evaluation rather than per-inference monitoring.

\FloatBarrier
\subsection{Horizontal Scaling}
\label{sec:scaling_results}

\begin{figure}[htbp]
    \centering
    \includegraphics[width=0.95\linewidth]{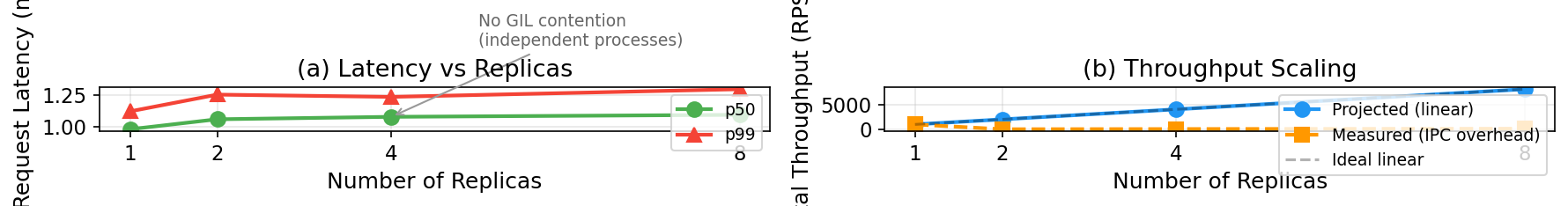}
    \caption{Horizontal scaling. \textbf{(a)}~Per-request latency remains constant across replica counts. \textbf{(b)}~Projected throughput scales linearly; observed multiprocessing throughput is limited by IPC serialization, not service compute.}
    \label{fig:scaling}
\end{figure}

We use multiprocessing (not threads) to avoid Python GIL artifacts. Per-request latency remains essentially constant at 0.98--1.10\,ms p50 regardless of replica count (1 to 8), confirming no resource interference between workers. A single replica achieves 1{,}020 RPS. \Figref{fig:scaling} shows the scaling behavior.

The low aggregate RPS in the multiprocessing simulation (41--151 RPS for 2--8 replicas) is an artifact of Python's \texttt{ProcessPoolExecutor} IPC serialization overhead (NumPy array pickling), not a scaling limitation. In a real Kubernetes deployment, each replica runs as an independent pod with its own HTTP server, and request dispatch happens via the Kubernetes Service load balancer with negligible overhead ($\sim$0.1\,ms). Based on the per-pod throughput of 1{,}020 RPS, projected aggregate throughput scales linearly: 2{,}040 RPS at 2 replicas, 4{,}080 at 4, and 8{,}160 at 8.

\FloatBarrier
\subsection{Pipeline Orchestrator}
\label{sec:orchestrator_results}

\begin{table}[htbp]
\centering
\caption{Pipeline orchestrator validation. All mechanisms---DAG execution, retry, idempotency, and backpressure---are empirically verified.}
\label{tab:orchestrator}
\begin{tabular}{@{}llr@{}}
\toprule
\textbf{Test} & \textbf{Result} & \textbf{Time (ms)} \\
\midrule
Full pipeline (5 nodes) & Success & 22.0 \\
Quick check (2 nodes) & Success & 6.8 \\
Retry behavior & 2 retries before success & --- \\
Idempotency cache & $4.86 \rightarrow 0.03$\,ms ($157\times$) & --- \\
Backpressure (max=1) & Sequential execution & 11.7 \\
Backpressure (max=2) & 2-way parallelism & 10.2 \\
Backpressure (max=4) & Full parallelism & 8.6 \\
\bottomrule
\end{tabular}
\end{table}

\begin{figure}[htbp]
    \centering
    \includegraphics[width=0.95\linewidth]{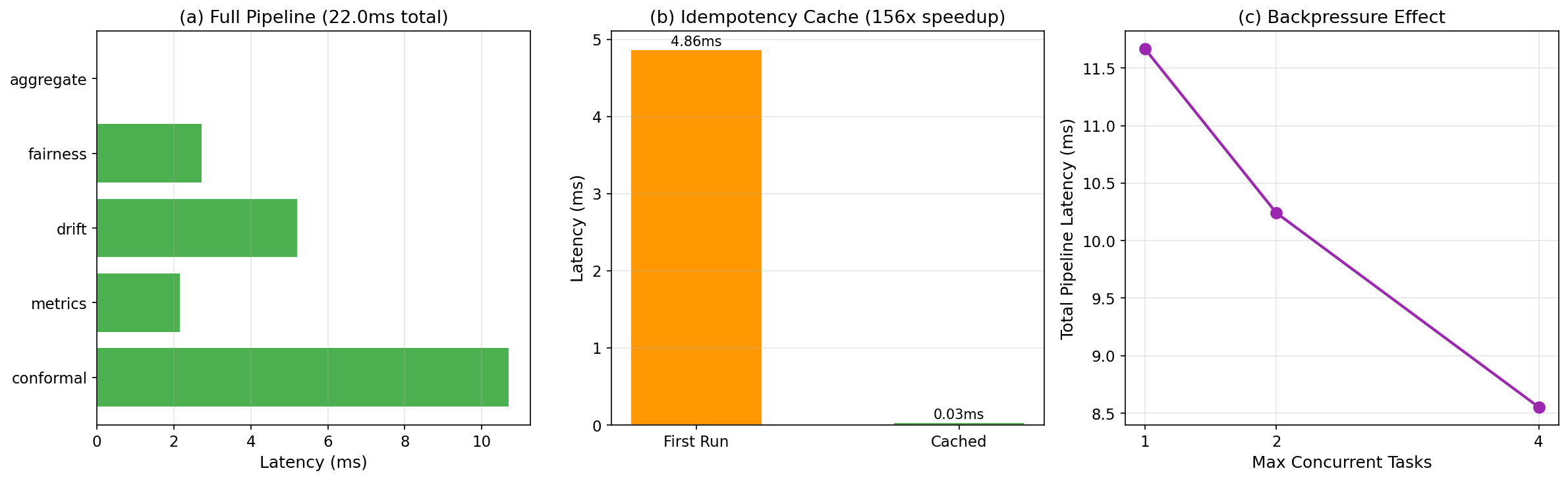}
    \caption{Pipeline orchestrator validation. \textbf{(a)}~Full pipeline DAG execution (22.0\,ms total). \textbf{(b)}~Idempotency cache provides $157\times$ speedup. \textbf{(c)}~Backpressure via semaphore controls concurrency.}
    \label{fig:orchestrator}
\end{figure}

\Tabref{tab:orchestrator} validates the pipeline orchestrator (\Figref{fig:orchestrator}). The full five-node evaluation DAG (conformal, metrics, drift, fairness, aggregate) executes in 22.0\,ms.

 The individual node latencies are: conformal 10.72\,ms, metrics 2.17\,ms, drift 5.21\,ms, fairness 2.74\,ms, and aggregate 0.02\,ms. Retry behavior is verified with 2 retries before success on a transient failure. The idempotency cache provides a $157\times$ speedup on cache hits (4.86\,ms $\rightarrow$ 0.03\,ms). Backpressure via semaphore correctly limits concurrency: total latency decreases from 11.7\,ms (sequential, max\_concurrent=1) to 8.6\,ms (parallel, max\_concurrent=4).

\FloatBarrier
\subsection{Comparison with Existing Tools}
\label{sec:comparison}

\begin{figure}[htbp]
    \centering
    \includegraphics[width=0.95\linewidth]{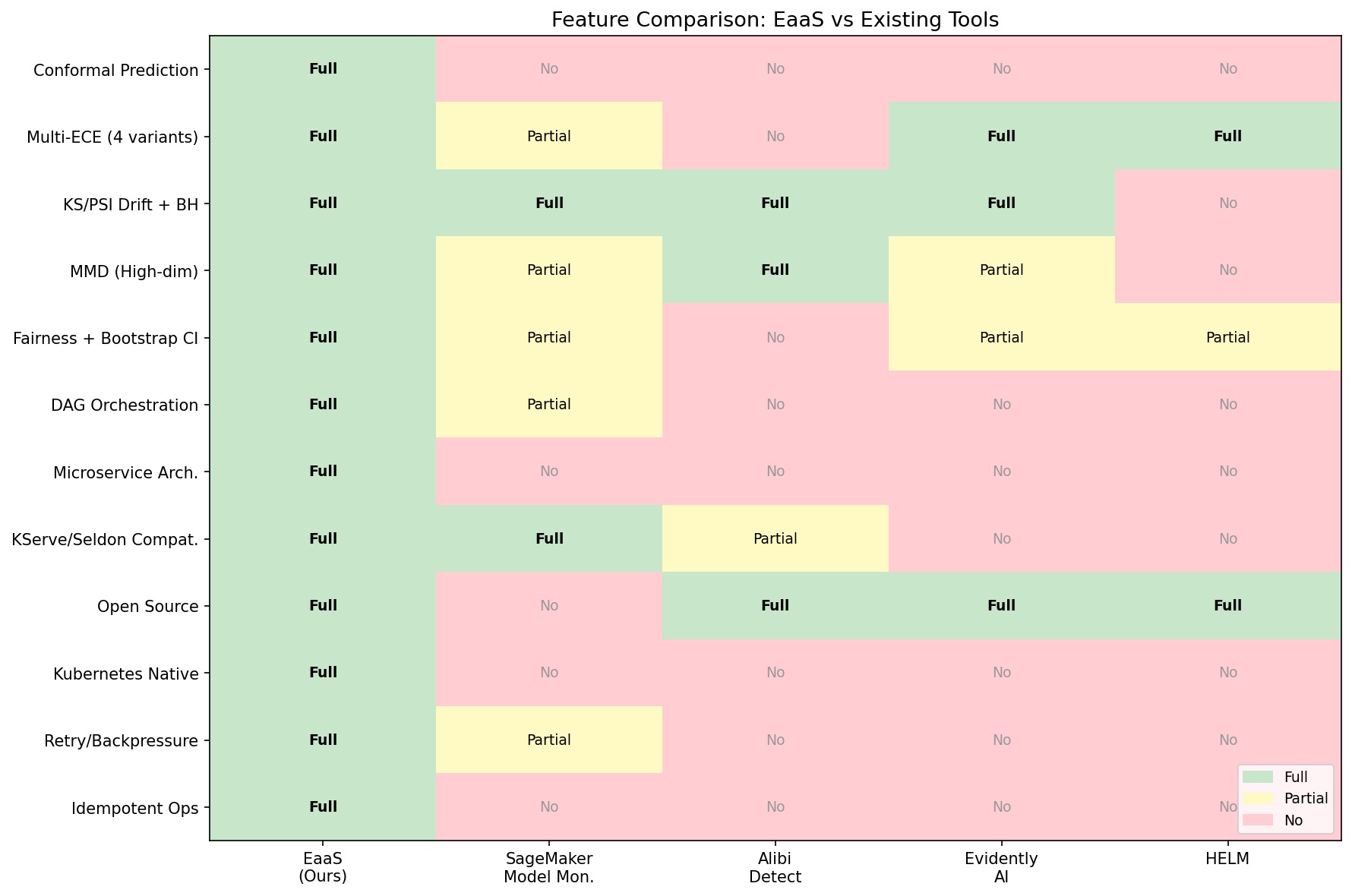}
    \caption{Feature comparison heatmap of \eaas against existing monitoring and evaluation tools across 12 capability dimensions.}
    \label{fig:comparison}
\end{figure}

\begin{table}[htbp]
\centering
\caption{Feature comparison of \eaas against existing monitoring and evaluation tools. \ding{51} = full support, $\sim$ = partial support, \ding{55} = no support.}
\label{tab:comparison}
\resizebox{\textwidth}{!}{%
\begin{tabular}{@{}lccccc@{}}
\toprule
\textbf{Feature} & \textbf{\eaas (Ours)} & \textbf{\sagemaker} & \textbf{\alibi} & \textbf{\evidently} & \textbf{\helm} \\
\midrule
Conformal Prediction & \ding{51} & \ding{55} & \ding{55} & \ding{55} & \ding{55} \\
Multi-ECE (4 variants) & \ding{51} & $\sim$ & \ding{55} & \ding{51} & \ding{51} \\
KS/PSI + BH correction & \ding{51} & \ding{51} & \ding{51} & \ding{51} & \ding{55} \\
\mmd (high-dim, approx.) & \ding{51} & $\sim$ & \ding{51} & $\sim$ & \ding{55} \\
Fairness + Bootstrap CI & \ding{51} & $\sim$ & \ding{55} & $\sim$ & $\sim$ \\
DAG Orchestration & \ding{51} & $\sim$ & \ding{55} & \ding{55} & \ding{55} \\
Microservice Architecture & \ding{51} & \ding{55} & \ding{55} & \ding{55} & \ding{55} \\
KServe Compatible & \ding{51} & \ding{51} & $\sim$ & \ding{55} & \ding{55} \\
Open Source & \ding{51} & \ding{55} & \ding{51} & \ding{51} & \ding{51} \\
Kubernetes Native & \ding{51} & \ding{55} & \ding{55} & \ding{55} & \ding{55} \\
Retry/Backpressure & \ding{51} & $\sim$ & \ding{55} & \ding{55} & \ding{55} \\
Idempotent Operations & \ding{51} & \ding{55} & \ding{55} & \ding{55} & \ding{55} \\
\bottomrule
\end{tabular}%
}
\end{table}

\Figref{fig:comparison} provides a visual heatmap of the feature comparison. \Tabref{tab:comparison} highlights four key differentiators of \eaas: (1)~conformal-prediction-as-a-service, which none of the open-source tools reviewed offers; (2)~microservice decomposition, versus library-based (\alibi, \evidently) or monolithic managed (\sagemaker) architectures; (3)~DAG orchestration with verified retry, idempotency, and backpressure; and (4)~approximate \mmd via \rff for practical high-dimensional drift detection as a scalable service. We note that the comparison is necessarily coarse-grained: ``full support'' means the feature is available out of the box as documented, ``partial'' means achievable with additional configuration or external integration (\eg \alibi can be containerized and served behind a REST API, though it does not ship as a Kubernetes-native service), and ``no support'' means the capability is absent from the tool's scope. Commercial platforms such as Arize and WhyLabs provide complementary production monitoring capabilities but are excluded from this table as they are not open-source and thus not directly comparable on implementation-level features.

\section{Discussion}
\label{sec:discussion}

\para{Conformal prediction in production.}
Our results confirm that the finite-sample correction $\lceil(n{+}1)(1{-}\alpha)\rceil/n$ delivers on its theoretical promise. The guarantee is marginal: $1 - \alpha \leq P(Y \in C(X)) \leq 1 - \alpha + 1/(n_{\text{cal}}{+}1)$, bounding coverage \emph{overshoot} due to the discrete calibration set. On synthetic data with $n_{\text{test}} = 5{,}000$, empirical coverage deviates from target by at most 1.3\%, consistent with test-set sampling variability. Crucially, the multi-split analysis (P1A) demonstrates that this is not an artifact of a single favorable partition: across 50 random calibration/test splits, mean coverage tracks the $1{-}\alpha$ target at all significance levels, with Wilson 95\% CIs consistently containing the target. On real \gptmini \mmlu logits, empirical coverage is consistent with the marginal conformal guarantee despite substantial miscalibration (debiased ECE~$\approx 0.55$). The logprob imputation analysis (P1B) further confirms robustness: all four MMLU answer tokens were present in the top-20 logprobs, and simulated imputation at 10\% has negligible impact ($\leq 1.3\%$ coverage deviation). At higher imputation fractions, the system conservatively expands prediction sets, maintaining the coverage guarantee at the cost of informativeness---a safe failure mode for production monitoring.

\para{Choosing ECE estimators.}
The four ECE variants serve complementary purposes. Fixed-bin ECE is widely used (\eg \helm uses 10 equal-width bins) but varies by up to 0.02 across bin counts. Adaptive-bin ECE offers robustness to bin configuration, making it suitable for automated monitoring. Debiased ECE provides the most accurate estimate by removing approximately 0.005 of finite-sample bias. We recommend debiased ECE for reporting, adaptive ECE for automated monitoring, and fixed-bin ECE only for comparability with prior work.

\para{Two-tier drift detection.}
Our results support a two-tier strategy: a \emph{fast path} using per-feature KS/PSI tests (sub-10\,ms for $\leq 10$ features, suitable for per-batch monitoring) and a \emph{deep path} using approximate \mmd ($\sim$500\,ms, suitable for periodic evaluation). The KS test with BH correction correctly identifies all 5 drifted features at magnitude $\geq 0.2$ with zero false positives, while \mmd detects multivariate distributional shifts that per-feature tests cannot capture. The \rff-\mmd ablations (P2A) validate three key operational choices: (i)~$n_{\text{rff}} = 500$ yields low approximation variance (std $\approx 0.001$) and drift-detection decisions consistent with exact \mmd; (ii)~the median heuristic for bandwidth yields Type~I error $\leq 0.085$ with full power at $\sigma \geq 0.5\times\sigma_{\text{median}}$, while under-smoothing ($0.25\times$) degrades power to 0.065; and (iii)~200 permutations provide well-calibrated $p$-values (rejection rate 0.07 at $\alpha{=}0.05$), while 50 permutations inflate Type~I error due to discretization.

\para{Pipeline orchestrator trade-offs.}
The lightweight orchestrator ($\sim$300 LOC) is purpose-built for evaluation pipelines, in contrast to general-purpose workflow engines such as Argo Workflows or Kubeflow Pipelines that require heavy Kubernetes CRD installations. Its verified retry (2 retries before success), idempotency ($157\times$ speedup), and backpressure (semaphore-based concurrency control) address the specific reliability needs of evaluation workloads. For complex MLOps workflows spanning training, deployment, and evaluation, Argo or Temporal may serve as the outer orchestrator, invoking \eaas pipelines as a composable evaluation step.

\para{NIST AI RMF alignment.}
The \eaas architecture maps to three core NIST AI Risk Management Framework functions \citep{nist2023rmf}: \emph{Measure} (continuous evaluation via Metrics Assessment and Conformal Prediction services, including pre-deployment evaluation DAGs), \emph{Manage} (drift detection triggers re-evaluation; fairness alerts with configurable thresholds and severity levels), and \emph{Govern} (audit trail via Result API with timestamped evaluations, reproducible evaluation via configurable DAG pipelines). This alignment provides a compliance pathway for organizations adopting the NIST framework.

\para{Privacy considerations.}
The minimum-sample threshold (default 30) serves a dual purpose: preventing noisy metrics for small subgroups and providing a first line of defense against inference attacks on rare populations. Bootstrap CI width naturally increases for small groups, providing built-in uncertainty communication. The Result API stores aggregate metrics only---no individual predictions or protected attributes---supporting privacy-by-design principles.

\para{Limitations.}
We identify seven limitations.
(1)~Scaling results use multiprocessing simulation rather than actual multi-pod Kubernetes deployment; per-request latency is representative (independent processes), but network overhead ($\sim$0.1--1\,ms per pod hop) and HPA scale-out times (typically 30--60\,s) are not empirically measured.
(2)~The OpenAI logprobs API returns only the top-20 tokens; however, our analysis (P1B) found that all four MMLU answer tokens were present in the top-20 for all 500 questions, and simulated imputation at realistic fractions has bounded impact on coverage.
(3)~We validated on \gptmini only; multi-model comparison would strengthen the CP results, though multi-split analysis (P1A) confirms robustness across 50 random partitions.
(4)~\rff-\mmd with 200 permutations on 768-dimensional embeddings takes $\sim$500\,ms, limiting it to batch evaluation. Streaming approaches such as MMDEW \citep{kalinke2022mmdew} or cached \rff projections could reduce this.
(5)~Fairness monitoring is now validated on real \adultincome data (P2B), but with an education-based proxy classifier; evaluation with a production ML model would further strengthen external validity.
(6)~The current architecture focuses on classification evaluation and does not address generative evaluation (LLM-as-judge, reference-free metrics).
(7)~Bootstrap confidence intervals assume independent samples; temporal correlations in production data could affect coverage. Block bootstrap or moving-window estimates may be more appropriate for streaming production data.

\section{Conclusion}
\label{sec:conclusion}

We presented \eaas, a cloud-native reference architecture that operationalizes mathematically rigorous AI evaluation methods as six stateless Kubernetes microservices. Our empirical validation demonstrates six key findings:
(1)~conformal prediction with the finite-sample-corrected \aps score produces empirical coverage consistent with the marginal $1{-}\alpha$ guarantee on both synthetic data and real \gptmini \mmlu logits, even under substantial miscalibration (ECE~$\approx 0.56$), with robustness confirmed across 50 random calibration/test splits;
(2)~the conformal prediction and calibration services achieve sub-2\,ms p99 latency at batch size 100 in single-process benchmarks (sub-10\,ms at batch size 500), while \rff-\mmd requires $\sim$500\,ms and is suited for periodic batch monitoring;
(3)~approximate \mmd via Random Fourier Features produces drift-detection decisions consistent with exact \mmd on 768-dimensional embeddings, with Type~I error and power well-calibrated across bandwidth and permutation count settings;
(4)~the pipeline orchestrator executes a full five-node evaluation DAG in 22\,ms with verified retry, idempotency ($157\times$ cache speedup), and backpressure control;
(5)~fairness monitoring on real UCI \adultincome data correctly identifies significant racial disparities (DP gap 0.330) with bootstrap CIs and stable alerts across sequential batches; and
(6)~among the open-source tools reviewed, \eaas is the only platform that combines conformal-prediction-as-a-service, microservice decomposition, and DAG-based orchestration.

\para{Future work.}
We identify three high-priority directions. First, deployment on a multi-node Kubernetes cluster (GKE/EKS) with actual HPA autoscaling and load testing to validate real scaling curves and measure inter-pod network overhead. Second, integration of streaming drift detection via MMDEW \citep{kalinke2022mmdew} exponential windows for online change detection with $O(\log^2 t)$ per-observation complexity. Third, extension to generative evaluation through LLM-as-judge services, BERTScore, and semantic equivalence scoring---operationalizing these as additional composable nodes within the \eaas evaluation DAG.

\clearpage
\bibliographystyle{plainnat}

\newpage
\appendix
\section{API Specification}
\label{sec:api}

Each \eaas service exposes a FastAPI auto-generated OpenAPI specification at \texttt{GET /docs}. We summarize the key endpoints below.

\para{Conformal Prediction Service (port 8001).}
\texttt{POST /calibrate} accepts softmax probabilities, labels, significance level $\alpha$, method (\aps or simple), and a randomize flag, returning the calibration threshold $\hat{q}$ and adjusted quantile level.
\texttt{POST /predict} generates prediction sets given softmax probabilities and the calibration parameters.

\para{Metrics Assessment Service (port 8002).}
\texttt{POST /evaluate} computes all calibration metrics (fixed/adaptive/classwise/debiased ECE, Brier score, log loss, accuracy) given softmax probabilities, labels, and bin count.

\para{Drift Detection Service (port 8003).}
\texttt{POST /detect/univariate} runs KS and PSI tests with configurable multiple-testing correction (BH, Bonferroni, or none) across features, returning per-feature p-values and the set of drifted features.
\texttt{POST /detect/mmd} computes approximate \mmd via \rff with configurable $n_{\text{rff}}$, bandwidth, and number of permutations, with optional exact \mmd comparison.

\para{Fairness Monitoring Service (port 8004).}
\texttt{POST /assess} performs full fairness assessment including demographic parity, equalized odds, bootstrap confidence intervals, minimum-sample filtering, and alert generation with configurable thresholds and severity levels.

\para{Pipeline Orchestrator (port 8005).}
\texttt{GET /pipelines} lists available DAG pipeline configurations (\eg \texttt{full\_evaluation}, \texttt{quick\_check}).
\texttt{GET /pipelines/\{name\}} retrieves a specific pipeline definition with node dependencies and configurations.

\para{Result API (port 8006).}
\texttt{POST /results} persists evaluation results with timestamps, model identifiers, and metadata.
\texttt{GET /results/\{id\}} retrieves a result by identifier.
\texttt{GET /results} supports listing and querying results with filtering by model and service.

All services expose a \texttt{GET /health} endpoint for Kubernetes liveness and readiness probes.

\section{Orchestrator Comparison}
\label{sec:orchestrator_comparison}

\begin{table}[h]
\centering
\caption{Comparison of the \eaas orchestrator with general-purpose workflow engines.}
\label{tab:orchestrator_comparison}
\resizebox{\textwidth}{!}{%
\begin{tabular}{@{}lcccc@{}}
\toprule
\textbf{Feature} & \textbf{\eaas Orchestrator} & \textbf{Argo Workflows} & \textbf{Kubeflow Pipelines} & \textbf{Temporal} \\
\midrule
DAG support & \ding{51} & \ding{51} & \ding{51} & \ding{51} \\
Retry + backoff & \ding{51} (verified) & \ding{51} & \ding{51} & \ding{51} \\
Backpressure & \ding{51} (semaphore) & Container-level & Container-level & Activity-level \\
Idempotency & \ding{51} ($157\times$) & Manual & Manual & Built-in \\
Evaluation-specific & \ding{51} & \ding{55} & \ding{55} & \ding{55} \\
Lightweight & \ding{51} ($<$500 LOC) & Heavy (CRDs) & Heavy (K8s deps) & Medium \\
\bottomrule
\end{tabular}%
}
\end{table}

\section{NIST AI RMF Mapping}
\label{sec:nist_mapping}

\begin{table}[h]
\centering
\caption{Mapping of \eaas capabilities to NIST AI Risk Management Framework functions \citep{nist2023rmf}.}
\label{tab:nist}
\resizebox{\textwidth}{!}{%
\begin{tabular}{@{}llll@{}}
\toprule
\textbf{RMF Function} & \textbf{Category} & \textbf{\eaas Capability} & \textbf{Service} \\
\midrule
Measure & 2.5 Effectiveness & Accuracy, ECE, Brier score evaluation & Metrics Assessment \\
Measure & 2.6 Pre-deployment & Full evaluation DAG before model promotion & Pipeline Orchestrator \\
Measure & 2.7 Uncertainty & Conformal prediction sets with coverage guarantees & Conformal Prediction \\
Manage & 3.1 Risk monitoring & Drift detection triggers re-evaluation & Drift Detection \\
Manage & 3.2 Risk response & Fairness alerts with configurable thresholds & Fairness Monitoring \\
Govern & 1.1 Governance & API documentation, configurable pipelines & All services \\
Govern & 1.5 Audit trail & Evaluation results persisted with timestamps & Result API \\
\bottomrule
\end{tabular}%
}
\end{table}

\end{document}